\documentclass[runningheads,orivec]{llncs}
\usepackage{graphicx}
\usepackage{subcaption}
\usepackage{capt-of}
\usepackage{amsmath,amssymb} %
\usepackage{color}
\usepackage{float}
\usepackage{array}
\usepackage{subfiles}
\usepackage{comment}
\usepackage[pagebackref=true,breaklinks=true,colorlinks,bookmarks=false]{hyperref}
\usepackage{booktabs}
\usepackage{cleveref}
\usepackage[export]{adjustbox}

\newif\ifanonymous

\newif\ifarxiv
\arxivtrue %

\ifarxiv
    \usepackage[width=122mm,left=12mm,paperwidth=146mm,height=193mm,top=12mm,paperheight=217mm]{geometry}
\fi

\ifanonymous
    \usepackage{ruler}
    \usepackage[width=122mm,left=12mm,paperwidth=146mm,height=193mm,top=12mm,paperheight=217mm]{geometry}
\fi


\definecolor{orange}{rgb}{1.0,0.65,0.0}
\definecolor{mygreen}{rgb}{0.05,0.5,0.35}
\newif\ifcomments
\commentstrue %
\ifcomments
    \newcommand{\roni}[1]{\textcolor{blue}{[S: #1]}}
    \newcommand{\eli}[1]{\textcolor{mygreen}{[E: #1]}}
    \newcommand{\ira}[1]{\textcolor{green}{[I: #1]}}
    \newcommand{\roy}[1]{\textcolor{cyan}{[R: #1]}}
    \newcommand{\ohad}[1]{\textcolor{orange}{[O: #1]}}
    
\else
    \providecommand{\roni}[1]{}
    \providecommand{\eli}[1]{}
    \providecommand{\ira}[1]{}
    \providecommand{\roy}[1]{}
    \providecommand{\ohad}[1]{}
\fi

\begin{document}

\pagestyle{headings}
\mainmatter
\def\ECCVSubNumber{88}  %

\title{Lifespan Age Transformation Synthesis}

\ifanonymous
    \titlerunning{ECCV-20 submission ID \ECCVSubNumber} 
    \authorrunning{ECCV-20 submission ID \ECCVSubNumber} 
    \author{Anonymous ECCV submission}
    \institute{Paper ID \ECCVSubNumber}
\else
    \titlerunning{Lifespan Age Transformation Synthesis}

    \author{Roy Or-El\inst{1} \and
    Soumyadip Sengupta\inst{1} \and
    Ohad Fried\inst{2} \and \\
    Eli Shechtman\inst{3} \and
    Ira Kemelmacher-Shlizerman\inst{1}}
    
    \authorrunning{Or-El R. et al.}
    
    \institute{\textsuperscript{1}University of Washington \quad
    \textsuperscript{2}Stanford University \quad
    \textsuperscript{3}Adobe Research}
\fi

\maketitle

\begin{abstract}
We address the problem of single photo age progression and regression---the prediction of how a person might look in the future, or how they looked in the past. Most existing aging methods are limited to changing the texture, overlooking transformations in head shape that occur during the human aging and growth process. This limits the applicability of previous methods to aging of adults to slightly older adults, and application of those methods to photos of children does not produce quality results. We propose a novel \textit{multi-domain image-to-image generative adversarial network} architecture, whose learned latent space models a continuous bi-directional aging process. The network is trained on the FFHQ dataset, which we labeled for ages, gender, and semantic segmentation. Fixed age classes are used as anchors to approximate  continuous age transformation. Our framework can predict a full head portrait for ages 0--70 from a single photo, modifying both texture and shape of the head. We demonstrate results on a wide variety of photos and datasets, and show significant improvement over the state of the art. 
\end{abstract}

\begin{figure*}[t]
\centering
\includegraphics[width=\textwidth]{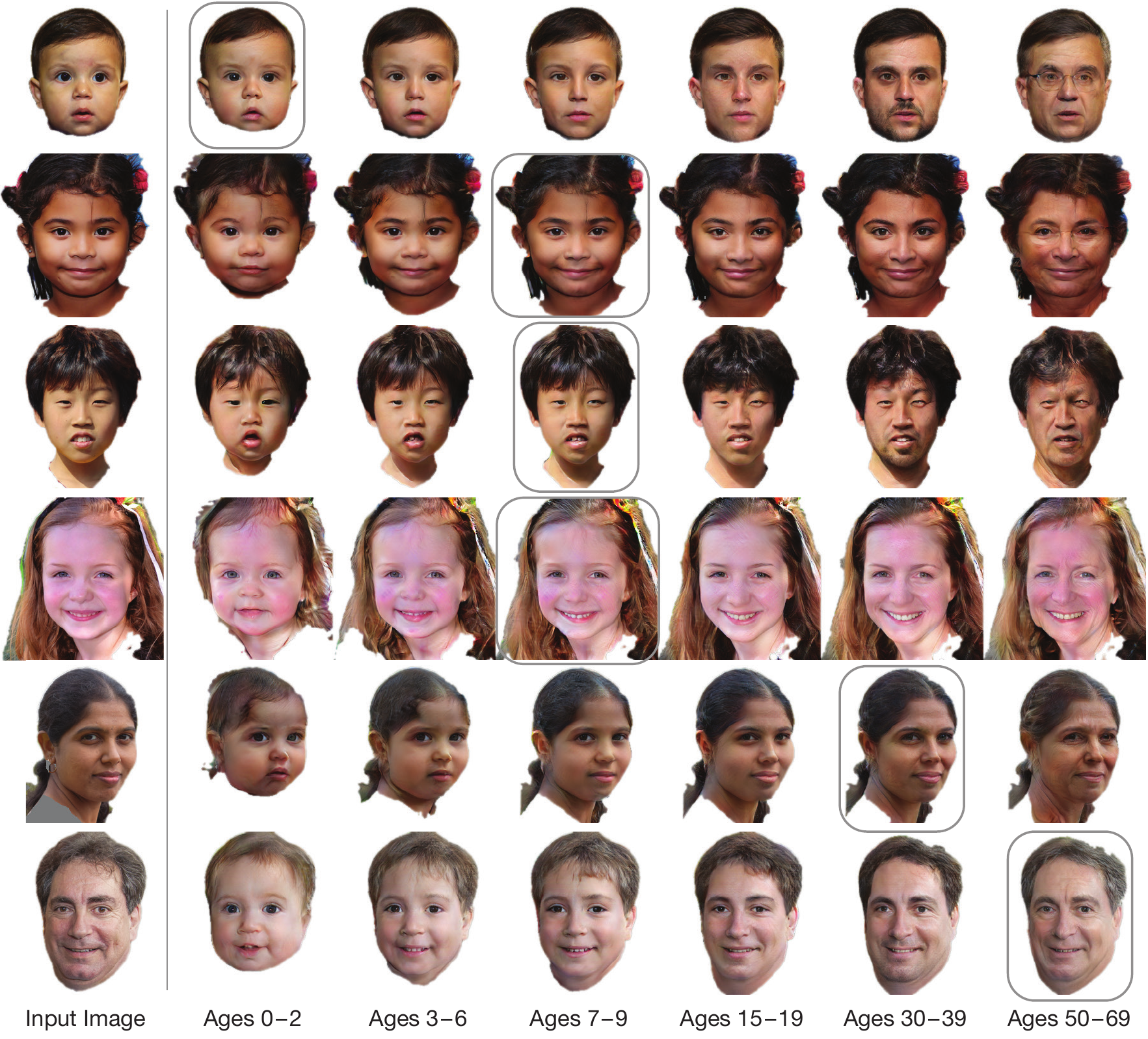}
\captionof{figure}{Given a single portrait photo (left), our generative adversarial network predicts full head bi-directional age transformation. Note the diversity of the input photos spanning ethnicity, age (baby, young, adult, senior), gender, and facial expression. Gray border marks the input class, columns 2-7 are all synthesized by our method.}
\label{fig:teaser}
\end{figure*}

\section{Introduction}
\label{sec:intro}

Age transformation is a problem of synthesizing a person's appearance in a different age while preserving their identity. Once the age gap between the input and the desired output is significant, e.g., going from 1 to 15 year old, the problem becomes highly challenging due to pronounced changes in head shape as well as facial texture. 
Solving for shape and texture together remains an open problem. Particularly, if the method is required to create a \textit{lifespan} of transformations, i.e., for any given input age, the method should synthesize a full span of 0--70 ages (rather than binary young-to-old transformations). In this paper, we aim to enable exactly that---lifespan of transformations from a single portrait. 

State of the art methods~\cite{wang2016recurrent,antipov2017face,Zhang_2017_CVPR,Duong_2017_ICCV,yang2017learning,wang2018face,he2019s2gan} focus on either minor age gaps, or mostly on adults to elderly progression, as a large part of the aging transformation for adults lies in the texture (rather than shape), e.g., adding wrinkles. The method of Kemelmacher-Shlizerman et al.~\cite{Kemelmacher-Shlizerman_2014_CVPR}  allows substantial age transformations but it can be applied only on a cropped face area, rather than a full head, and cannot be  modified to allow backward age prediction (adult to child) due to optical-flow-based  nature of the method. Apps like FaceApp allow considerable transitions from adult to child and vice versa, but similar to state of the art methods they focus on texture, not shape, and thus produce sub-par results, in addition to focusing only on the binary case of two ambiguous age classes (``young'', ``old''). 

Theoretically, since time is a continuous variable, lifespan age transformation, e.g., 0--70 synthesis, should be modeled as a continuous process. However, it can be very difficult to learn without large datasets of identity-specific ground truth (same person captured over their lifespan). Therefore, we approximate this continuous transformation by representing age with a fixed number of anchor classes in a multi-domain transfer setting. We represent age with six anchor classes: three for children ages 0--2, 3--6, 7--9, one for young people 15--19, one for adults 30--39,  and one for 50--69. Those classes are designed to learn geometric transformation in ages where most prominent shape changes occur, while covering the full span of ages in the latent space.
 
To that end, we propose a new multi-domain image-to-image conditional GAN architecture (Fig.~\ref{fig:arch}). Our main encoder---the identity encoder---encodes the input image to extract features associated with the person's identity. Next, unlike other multi-domain approaches, the various age domains are each represented by a unique distribution. Given a target age, it is assigned an age vector code sampled from the appropriate distribution. The age code is sent to a mapping network, that maps age codes into a unified, learned latent space. The resulting latent space approximates continuous age transformations.  Our decoder then fuses the learned latent age representation with the identity features via StyleGAN2's~\cite{Karras2019stylegan2} modulated convolutions.  

Disentanglement based domain transfer approaches such as  MUNIT~\cite{huang2018munit} and FUNIT~\cite{liu2019funit} can learn shape and texture deformation, e.g., transform cats to dogs. However, these methods cannot be directly applied to transform age, in a multi-domain setting, due to key limiting assumptions: MUNIT requires two generators per domain pair, thus training it for even 6 age classes will require 30 generators, trumping scalability. FUNIT requires an exemplar image of the target class and is not guaranteed to apply only age features from the exemplar, as other attributes like skin color, gender and ethnicity may also be transferred.

On the other hand, multi-domain transfer algorithms such as StarGAN~\cite{choi2017stargan} and STGAN~\cite{liu2019stgan} assume the domains to be distinct and encompassing contrasting facial attributes. Age domains are highly correlated however, and thus those algorithms struggle with the age transformation task. Methods like InterFaceGAN~\cite{shen2019interpreting} aim to address that via latent space traversal of an unconditionally trained GAN. However, navigating these paths to transform a person into a specific age is difficult, as the computed traversal path does not always preserve identity characteristics. In contrast, our proposed algorithm can transform shape and texture across a wide range of ages while still maintaining the person's identity and attributes.

Another limiting factor in modeling full lifespan age transformations is that existing face aging datasets contain a very limited amount of babies and children. To compensate for that, we labeled the FFHQ dataset~\cite{karras2019style} for gender and age via crowd-sourcing. In addition, for each image we extracted face semantic maps as well as head pose angles, glasses type and eye occlusion score.  

Qualitative and quantitative evaluations show that our method outperforms state-of-the-art aging algorithms as well as multi-domain transfer and latent space traversal methods applied on the face aging task. The key contributions of this paper are: 1) enabling both shape and texture transformations for lifespan age synthesis, 2) novel multi-domain image-to-image translation GAN architecture, 3) labelled FFHQ~\cite{karras2019style} dataset which we will share with the community.

We are aware of the potential ethical issues and potential bias such method can present. These issues are addressed in our Ethics and Bias statement in the supplementary material.

\section{Related Work}
\label{sec:related}

Early works in age progression have focused on building separate models for specific sub-effects of aging, e.g.,  wrinkles~\cite{wu1994plastic,boissieux2000simulation,bando2002simple},  cranio-facial growth~\cite{ramanathan2006modeling,ramanathan2008modeling}, and face sub-regions ~\cite{suo2010compositional,suo2012concatenational}. Complete face transformation was explored via calculating average faces of age clusters and transitioning between them~\cite{burt1995perception,rowland1995manipulating},  wavelet transformation~\cite{tiddeman2001prototyping}, dictionary learning~\cite{shu2015personalized}, factor analysis~\cite{yang2016face} and AAM face parameter fitting~\cite{lanitis2002toward}. Age progression of children was specifically the focus in ~\cite{Kemelmacher-Shlizerman_2014_CVPR}, where the aging process was modelled as cascaded flows between mean faces of pre-computed age clusters of eigenfaces. 

Recently, deep learning has become the predominant approach for facial aging. Wang \emph{et al.}~\cite{wang2016recurrent} replaced the cascaded flows from~\cite{Kemelmacher-Shlizerman_2014_CVPR} with a series of RNN forward passes. Zhang \emph{et al.}~\cite{Zhang_2017_CVPR} and Antipov \emph{et al.}~\cite{antipov2017face} proposed autoencoder GAN architecture where aging was performed by adding an age condition to the latent space. Duong \emph{et al.}~\cite{nhan2016longitudinal,Duong_2017_ICCV} introduced a cascade of restricted Boltzmann machines and ResNet based probabilistic generative models (respectively) to carry out the aging process between age groups. Yang \emph{et al.}~\cite{yang2017learning} proposed a GAN based architecture with a pyramidal discriminator over age detection features of the generated aged image. Liu \emph{et al.}~\cite{liu2017face} introduced additional age transition discriminator to supervise the aging transitions between the age clusters. Li \emph{et al.}~\cite{li2018global} fused the outputs from global and local patch generators to synthesize the aged face. Wang \emph{et al.}~\cite{wang2018face} added facial feature loss as well as age classification loss to enforce the output image to have the same identity while still progressing the age. Liu \emph{et al.}~\cite{Liu_2019_CVPR} added gender and race attributes to their GAN architecture to help avoid biases in the training data, they also propose a new wavelet based discriminator to improve image quality. He \emph{et al.}~\cite{he2019s2gan} encoded personalized aging basis and apply specific age transforms to create an age representation used to decode the aged face. The focus of most of those approaches was on aging adults to elderly (mostly texture changes). Our method is the first to propose a full lifespan aging, 0--70 years old. We refer the reader to these excellent surveys~\cite{duong2018longitudinal,fu2010age,ramanathan2009computational} for a broader overview of the advances in age progression over the years.

Recent success with generative adversarial networks~\cite{goodfellow2014generative} significantly improved image-to-image translations between two domains, with both paired~\cite{Isola_2017_CVPR} (Pix2Pix) and unpaired~\cite{Zhu_2017_ICCV} (CycleGAN) training data. More recent methods disentangled the image into style and content latent spaces, e.g., MUNIT~\cite{huang2018munit}, DRIT~\cite{lee2018diverse}, share the content space but create multiple disjoint style latent spaces. These methods are hard to scale to a large number of domains as they require training two generators per pair of domains. FUNIT~\cite{liu2019funit} used a single generator that disentangled the image into shared content and style latent spaces, however, it required an additional target image to explicitly encode the style. One may consider aging effects as ``style''. However when transferring style between two age domains, non-age related styles, like skin color, gender and ethnicity, might be transferred as well. Multi-domain transfer algorithms like StarGAN~\cite{choi2019starganv2} and STGAN~\cite{liu2019stgan} can edit multiple facial attributes, but those are assumed to be distinct and contrasting. StarGAN generalizes CycleGAN to map an input image into multiple domains using a single generator. STGAN uses a selective transfer units with encoder-decoder architecture to select and modify encoded features for attribute editing. These methods however are not designed to work on the age translation task, as aging domains are highly correlated and not distinct. Our proposed architecture enables translations between highly correlated domains, and obtains a continuous traversable age latent space while maintaining  identity and image quality.

\section{Algorithm}
\label{sec:algo}
\subsection{Overview}
\label{alg_over}

Our main goal is to design an algorithm that can learn the head shape deformation as well as appearance changes across a wide range of ages. Ideally, one would turn to supervised learning to tackle this problem. However, since this process is continuous in nature, it requires a large amount of aligned image pairs of the same person at different ages that will span all possible transitions. Unfortunately, there are no existing large-scale datasets that capture aging changes over more than several years, let alone an entire life span. Furthermore, small scale datasets like FGNET~\cite{lanitis2002toward} capture subjects in different poses, environments and lighting conditions, making supervised training very challenging. To this end, we turn to adversarial learning and leverage the recent progress in unpaired image-to-image translation GAN architectures~\cite{taigman2016unsupervised,Zhu_2017_ICCV,choi2017stargan,huang2018munit,lee2018diverse,liu2019funit}. We propose to approximate the continuous aging process with six anchor age classes which results in a multi-domain transfer problem.

We propose a novel generative adversarial network architecture that consists of a single conditional generator and a single discriminator. The conditional generator is responsible for transitions across age groups, and consists of three parts: identity encoder, a mapping network, and a decoder. 
We assume that while a person's appearance changes with age their identity remains fixed. Therefore, we encode age and identity in separate paths.

Each age group is represented by a unique pre-defined distribution. When given a target age, we sample from the respective age group distribution and assign it a vector age code. The age code is sent to a mapping network, that maps it into a learned unified age latent space. The resulting latent space approximates continuous age transformations. The input image is processed separately by the identity encoder to extract identity features. The decoder takes the mapping network output, a target age latent vector, and injects it to the identity features using modulated convolutions, originally proposed in StyleGAN2~\cite{Karras2019stylegan2}. During training, we use an additional age encoder to relate between real and generated images to the pre-defined distributions of their respective age class.

For transformation to an age not represented in our anchor classes, we calculate age latent codes for its two neighboring anchor classes and perform linear interpolation to get the desired age code as input to the decoder. 

\begin{figure}[t]
\centering
\includegraphics[width=\textwidth]{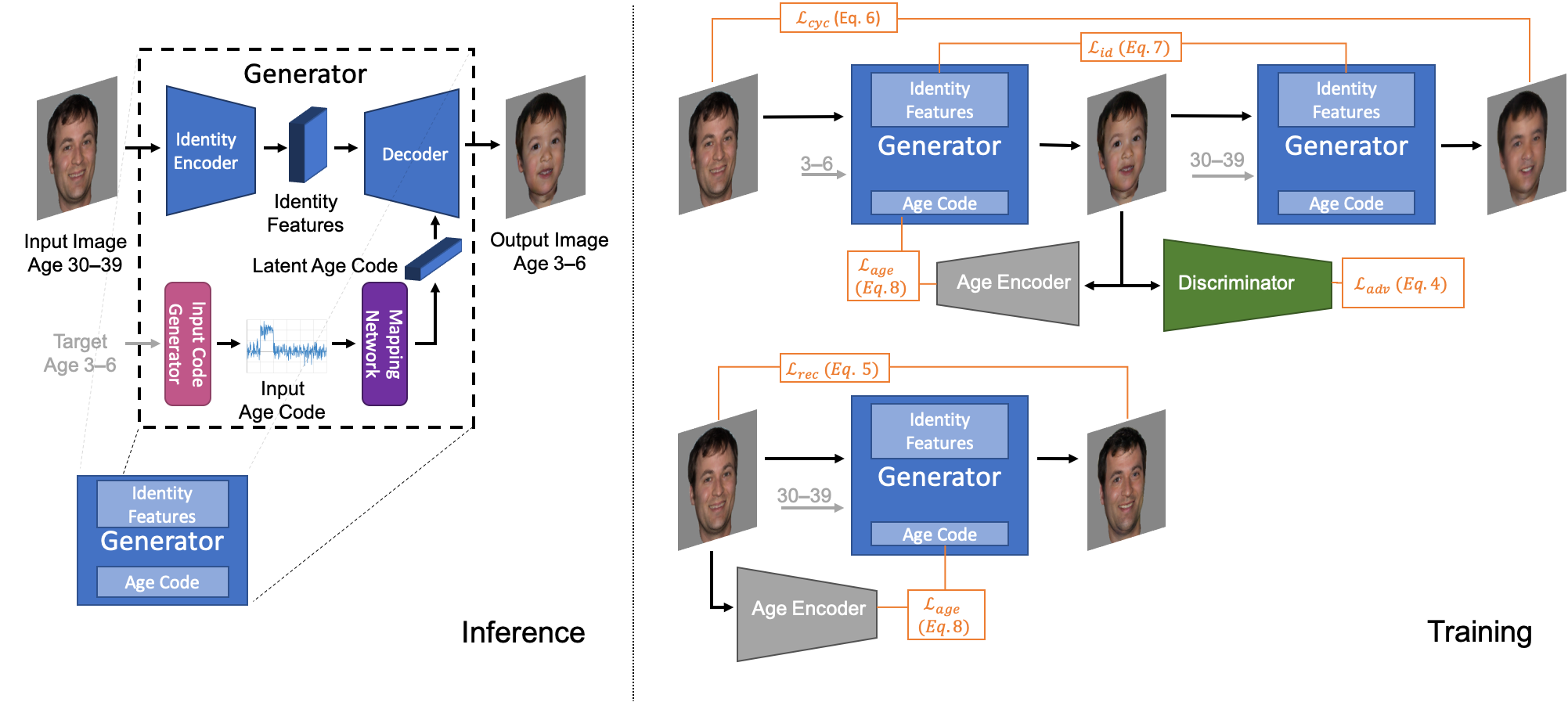}
\caption{Algorithm overview.}
\label{fig:arch}
\end{figure}

\subsection{Framework}
\label{alg_frame}
Our algorithm takes a facial image $x$ and a target age cluster as inputs. It then generates an output image of the same person at the desired age cluster. \figurename~\ref{fig:arch} shows the model architecture as well as the training scheme. 

As a pre-processing step, background and clothing items are removed from the image using its corresponding semantic mask, which is part of our dataset (see Sec.~\ref{sec:data} for details). Our age input space, $\mathcal{Z}$, is represented by a $50 \times n$ element vector where $n$ is the number of age classes. When the input age class is $i$, we generate a vector $z_i \in \mathcal{Z}$ as
\begin{equation}
    z_i = \mathbf{1}_i + v, \quad v \sim \mathcal{N}(0,0.2^2 \cdot\mathbf{I})
\end{equation}
where $\mathbf{1}_i$ is a $50 \cdot n$ element vector that contains ones on elements $50 \cdot i $ through $50 \cdot (i+1) -1$ and zeros elsewhere, and $\mathbf{I}$ is the identity matrix.
A single generator is used to generate all target ages. Our generator consists of an identity encoder, a latent mapping network and a decoder. During training we also use an age encoder to embed both real and generated images into the age latent space.

\textbf{Identity encoder}. The identity encoder $E_{id}$ takes an input image $x$ and extracts an identity features tensor $w_{id}$, where  $w_{id} = E_{id}(x)$. these features contain information about the image local structures and the general shape of the face which play a key role in generating the same identity. The identity encoder contains two downsampling layers followed by four residual blocks~\cite{he2016deep}.

\textbf{Mapping network}. The mapping network $M : \mathcal{Z} \to \mathcal{W}_{age}$ embeds an age input vector $z$ to the unified age latent space $\mathcal{W}_{age}$, $w_{age} = M(z)$, where $M$ is an 8 layer MLP network and $w_{age}$ is a 256 element latent vector. The mapping network learns an optimal age latent space that enables a smooth transition and interpolation between age clusters, needed for continuous age transformations.

\textbf{Decoder}. Our decoder takes an age latent code along with identity features and produces an output image $y = F(w_{id},w_{age})$. The identity features $w_{id}$ are processed by styled convolution blocks~\cite{karras2019style}. To reduce water droplet~\cite{Karras2019stylegan2} artifacts, we replace the AdaIN normalization layers~\cite{huang2017adain} with modulated convolution layers proposed in StyleGAN2~\cite{Karras2019stylegan2}. In addition, each modulated convolution layer is followed by a pixel norm layer~\cite{karras2017progressive} as we observed it further helps reducing these artifacts. We omit the noise injection in our implementation. Overall, we use four styled convolution layers to manipulate the identity code and two upsampling styled convolution layers to produce an image at the original size.

The overall generator mapping from an input image $x$ and an input target age vector $z_t$ to an output image $y$ is:
\begin{equation}
    y = G(x,z_t) = F(E_{id}(x),M(z_t)).
\end{equation}

\textbf{Age encoder}. The age encoder enforces a mapping of the input image $x$ into its correct location in the age vector space $\mathcal{Z}$. It produces an age vector $z_s = E_{age}(x)$ that corresponds to the source age cluster $s$ of the image $x$. The age encoder needs to capture more global data in order to encode the general appearance, regardless of the identity. To this end, we follow the architecture of MUNIT~\cite{huang2018munit}'s style encoder with four downsampling layers, followed by global averaging and a fully connected layer to produce an age vector. Note that the age encoder is not used for inference.

\textbf{Discriminator}. We use the StyleGAN discriminator~\cite{karras2019style} with minibatch standard deviation. We modify the last fully connected layer to have $n$ outputs in order to discriminate multiple classes as suggested by Liu \emph{et al.}~\cite{liu2019funit}. For a real image from class $i$, we only penalize the $i$-th output. Respectively, only the $j$-th output is penalized for a generated image of class $j$.

\subsection{Training Scheme}
\label{alg_train}
An overview of the training scheme can be seen in \figurename~\ref{fig:arch}. To compensate for imbalances between age clusters, in each training iteration, we first sample a source cluster $s$ and a target cluster $t$ ($t \neq s$). Then we sample an image from each class. We then perform three forward passes:
\begin{equation}
    y_{gen} = G(x,z_t),\quad y_{rec} = G(x_,z_s),\quad y_{cyc} = G(y_{gen},z_s). 
\end{equation}
Here, $y_{gen}$ is the generated image at target age $t$ and $y_{rec}$ is the reconstructed image at source age $s$. We also apply a cycle to reconstruct $y_{cyc}$ at source age $s$ from generated image $y_{gen}$ at age $t$. These passes provide us with all the necessary signals to minimize the following loss functions.

\textbf{Adversarial loss}. We use an adversarial loss conditioned on the source and target age cluster of the real and fake images respectively,
\begin{equation}
    \mathcal{L}_{adv}(G,D) = E_{x,s}[\log D_s(x)] + E_{x,t}[\log (1-D_t(y_{gen})],
\end{equation}
where $D_i$ is the $i$-th output of the discriminator, $s$ is the source age cluster of the real image and $t$ is the target cluster for the generated image.

\textbf{Self reconstruction loss}. This loss is used to force the generator to learn the identity translation. When the given target age cluster is the same as the source cluster, we minimize 
\begin{equation}
    \mathcal{L}_{rec}(G) = \|x - y_{rec}\|_1.
\end{equation}

\textbf{Cycle loss}. To help identity preservation as well as a consistent skin tone we employ the cycle consistency loss~\cite{Zhu_2017_ICCV}, 
\begin{equation}
    \mathcal{L}_{cyc}(G) = \|x - y_{cyc}\|_1.
\end{equation}

\textbf{Identity feature loss}. To make sure the generator keeps the identity of the person throughout the aging process, we minimize the $L1$ distance between the identity features of the original image and those of the generated image,
\begin{equation}
    \mathcal{L}_{id}(G) = \|E_{id}(x) - E_{id}(y_{gen})\|_1.    
\end{equation}

\textbf{Age vector loss}. We enforce a correct embedding of real and generated images to the input age space by penalizing the distance between the age encoder outputs and the age vectors $z_s, z_t$ that were sampled to generate outputs at the source and target age clusters respectively. The loss is defined as  
\begin{equation}
    \mathcal{L}_{age}(G) = \|E_{age}(x) - z_s\|_1 + \|E_{age}(y_{gen}) - z_t\|_1.    
\end{equation}

The overall optimization function is
\begin{equation}
    \begin{aligned}
        \underset{G}{\operatorname{min}}\ \underset{D}{\operatorname{max}}\ & \mathcal{L}_{adv}(G,D) \ + \lambda_{rec}\mathcal{L}_{rec}(G) + \\
        &\lambda_{cyc}\mathcal{L}_{cyc}(G) + \lambda_{id}\mathcal{L}_{id}(G) + \lambda_{age}\mathcal{L}_{age}(G).
    \end{aligned}
\end{equation}

\subsection{Implementation details}
We train 2 separate models, one for males and one for females. Each model was trained with a batch size of 12 for 400 epochs on 4 GeForce RTX 2080 Ti GPUs. We use the Adam optimizer~\cite{kingma2014adam} with $\beta_1=0$, $\beta_2=0.999$ and a learning rate of $10^{-3}$. The learning rate is decayed by 0.5 after 50 and 100 epochs. Similar to StyleGAN~\cite{karras2019style}, we apply the non-saturating adversarial loss~\cite{goodfellow2014generative} with $R1$ regularization~\cite{MeschederICML2018}. In addition, we also reduce the learning rate of the mapping network $M$ by a factor of $0.01$ and employ exponential moving average for the generator weights. We set $\lambda_{rec}=10$, $\lambda_{cyc}=10$, $\lambda_{id}=1$, $\lambda_{age}=1$. We refer the readers to the supplementary material for architecture details of each component in our framework. The \href{https://github.com/royorel/Lifespan_Age_Transformation_Synthesis}{code and pre-trained models} are publicly available.

\section{Dataset}
\label{sec:data}
\begin{figure}[t]
\centering
\includegraphics[width=\textwidth]{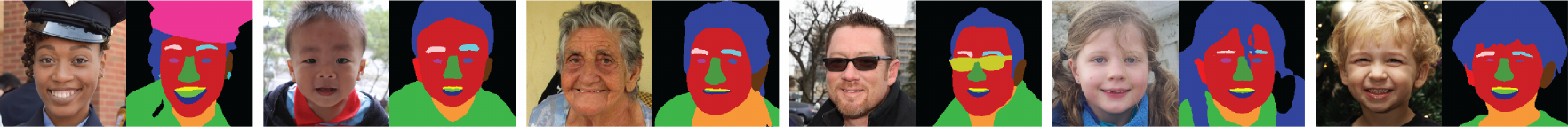}
\caption{FFHQ-Aging dataset. We label 70k images from FFHQ dataset \cite{karras2019style} for gender and age via crowd-sourcing. In addition, for each image we extracted face semantic maps as well as head pose angles, glasses type and eye occlusion score. }
\label{fig:data_samples}
\end{figure}

We introduce a new facial aging dataset based on images from FFHQ~\cite{karras2019style}, `FFHQ-Aging'. We used the Appen\footnote{\url{https://www.appen.com/}} crowd-sourcing platform to annotate gender and age cluster for all $70,000$ images on FFHQ, collecting 3 judgements for each image. We defined 10 age clusters that capture both geometric and appearance changes throughout a person's life: 0--2, 3--6, 7--9, 10--14, 15--19, 20--29, 30--39, 40--49, 50--69 and 70+. We trained a DeepLabV3~\cite{chen2017rethinking} network on the CelebAMask-HQ~\cite{CelebAMask-HQ} dataset and used the trained model to extract a 19 label face semantic maps for all 70K images. Finally, we used the Face++\footnote{\url{https://www.faceplusplus.com/}} platform to get the head pose angles, glasses type (none, normal, or dark) and left and right eye occlusion scores. We use the same alignment procedure as~\cite{karras2017progressive} with a slightly larger crop size (see supplementary for details). We generated our images and semantic maps at a resolution of 256x256 but the procedure is applicable to higher resolutions too. \figurename~\ref{fig:data_samples} shows sample image \& face semantics pairs from the dataset. There are 32,170 males and 37,830 females in the dataset, the age distribution per gender can be seen in the supplementary material. The  \href{https://github.com/royorel/FFHQ-Aging-Dataset}{dataset} is publicly available to the community.  

For the purpose of training our network, we assigned images 0--68,999, for training and images 69,000--69,999 for testing. Then, we pruned images with: gender confidence below $0.66$, age confidence below $0.6$,  head yaw angle greater than $40^{\circ}$, head pitch angle greater than $30^{\circ}$, dark glasses label, and eye occlusion score greater than 90 and 50 for eye pair. After pruning, we selected 6 age clusters to train on: 0--2, 3--6, 7--9, 15--19, 30--39, 50--69. This process resulted in 14,232 male and 14,066 female training images along with 198 male and 205 female images for testing. The pruned training set age distribution per gender is presented in the supplementary material.

\section{Evaluation}
\label{sec:eval}

\subsection{Comparison with Commercial Apps}
\begin{figure}[t]
\centering
    \begin{tabular}{p{0.116\textwidth}p{0.116\textwidth}p{0.116\textwidth}p{0.116\textwidth}p{0.116\textwidth}p{0.116\textwidth}p{0.116\textwidth}p{0.116\textwidth}}
    \includegraphics[width=0.115\textwidth]{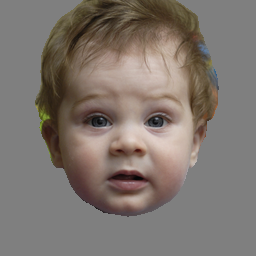}&
    \includegraphics[width=0.115\textwidth]{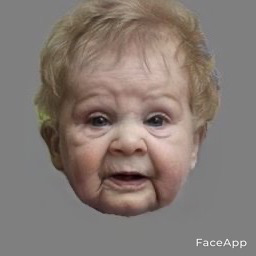}&
    \includegraphics[width=0.115\textwidth]{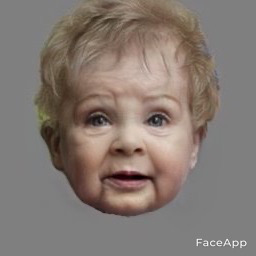}&
    \includegraphics[width=0.115\textwidth]{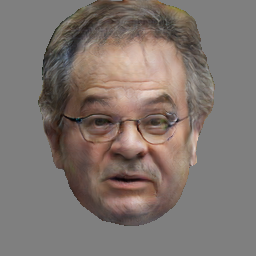}&
    \includegraphics[width=0.115\textwidth]{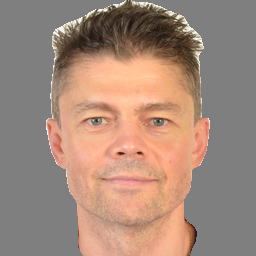}&
    \includegraphics[width=0.115\textwidth]{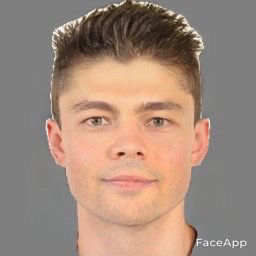}&
    \includegraphics[width=0.115\textwidth]{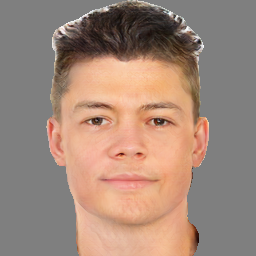}&
    \includegraphics[width=0.115\textwidth]{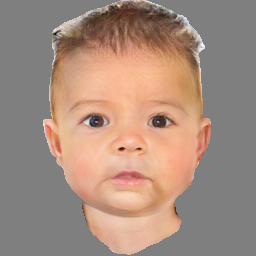}\tabularnewline
    \includegraphics[width=0.115\textwidth]{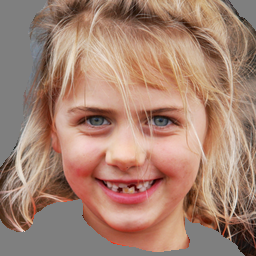}&
    \includegraphics[width=0.115\textwidth]{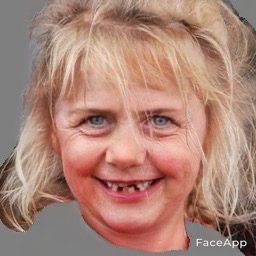}&
    \includegraphics[width=0.115\textwidth]{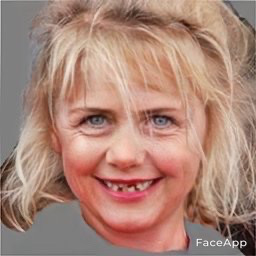}&
    \includegraphics[width=0.115\textwidth]{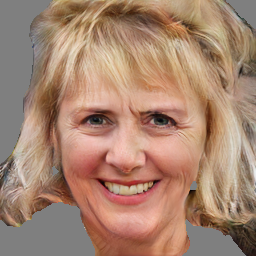}&
    \includegraphics[width=0.115\textwidth]{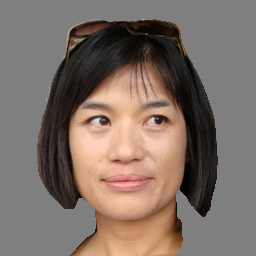}&
    \includegraphics[width=0.115\textwidth]{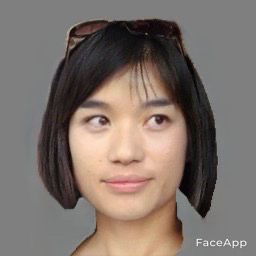}&
    \includegraphics[width=0.115\textwidth]{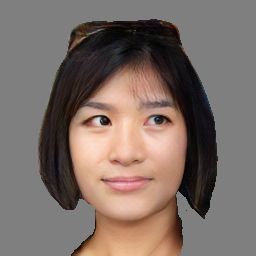}&
    \includegraphics[width=0.115\textwidth]{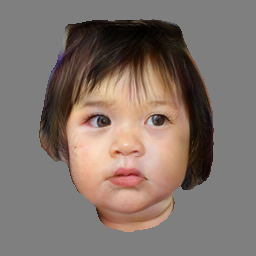}\tabularnewline
    \centering Input & \centering FaceApp Old & \centering FaceApp Cool Old & \centering Ours 50--69 & \centering Input & \centering FaceApp Young2 & \centering Ours 15--19 & \centering Ours 0--2
    \end{tabular}
    \caption{Comparison with FaceApp filters. Note that FaceApp cannot deform the shape of the head or generate extreme ages, e.g. 0--2.}
    \label{fig:faceapp}
\end{figure}

We perform a qualitative comparison with the outputs of FaceApp\footnote{\url{https://www.faceapp.com/}}. FaceApp provides binary facial aging filters to make people appear younger or older. \figurename~\ref{fig:faceapp} shows that although the FaceApp output image quality is high, it cannot perform shape transformation and is mostly limited to skin texture. For transformations to an older age, we applied both ``old'' and ``cool old'' filters available in FaceApp and compared against our output for 50--69 age range. For transformations to a younger age, we applied the ``young2'' filter which is roughly equivalent to our 15--19 class. We also show our outputs for the 0--2 class to demonstrate our algorithm's ability to learn head deformation. Even though FaceApp applies a dedicated filter for each transition, in contrast with our multi-domain generator, its age filters are still not transforming the shape of the head.

\subsection{Comparison with Age transformation methods}
We compare our algorithm to three state-of-the-art age transformation methods: IPCGAN~\cite{wang2018face}, Yang \emph{et al.}~\cite{yang2017learning}, referred as PyGAN, and S2GAN~\cite{he2019s2gan}. 

\textbf{Qualitative Evaluation.}
We compare with PyGAN and S2GAN on CACD dataset~\cite{chen14cross} %
on the images showcased by the authors in their papers. We train on FFHQ and test on CACD, while both PyGAN and S2GAN train on CACD dataset. Due to copyright issues with CACD images, we cannot present the comparison figures in this manuscript. We encourage the reader to review these images in the \href{https://grail.cs.washington.edu/projects/lifespan_age_transformation_synthesis/}{project's website}.
Even though PyGAN is trained with a different generator to produce each age cluster, our network is still able to achieve better photorealism for multiple output classes with a single generator. In comparison to S2GAN, our algorithm is able to create more pronounced wrinkles and facial features as the age progresses, all while spanning wider range of age transformations. 

\begin{figure*}[t!]
\centering
\includegraphics[width=0.98\textwidth]{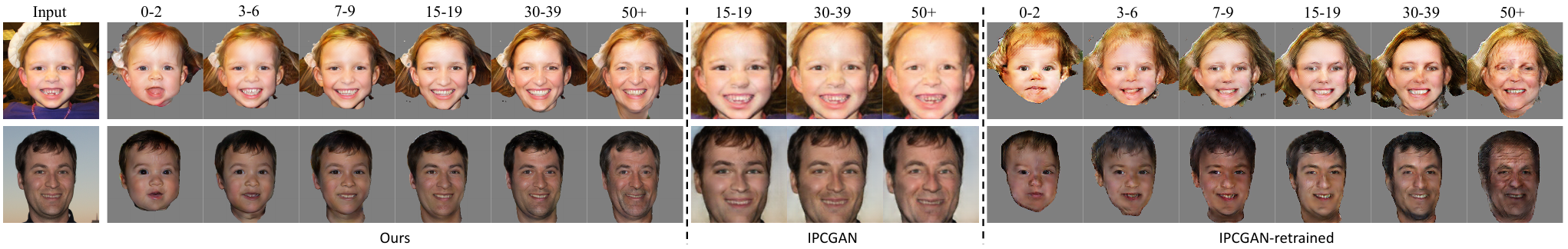}
\caption{Comparison w.r.t. IPCGAN~\cite{wang2018face} on the FFHQ-Aging dataset. Left: our method. Middle: IPCGAN trained on CACD. Right: IPCGAN trained on FFHQ-Aging. The proposed framework outperforms IPCGAN, producing sharper, more detailed and more realistic outputs across all age classes.}
\label{fig:ipcgan}
\end{figure*}

We also evaluate our performance w.r.t IPCGAN trained on both CACD \& FFHQ-Aging datasets in \figurename~\ref{fig:ipcgan}. Here, we use IPCGAN's publically available code and retrain their framework on FFHQ-Aging dataset for fair comparison (termed as `IPCGAN-retrained'). Our method outperforms both IPCGAN models in terms of image quality and shape deformation.

\textbf{User Study}
In addition, we performed a user study to evaluate PyGAN results vs. our results. In the study we measure: (a) how well does the method preserve the identity of the person in the photo, (b) how close is the perceived age to the target age, and (c) overall which result is better. Our hypothesis was that PyGAN will excel in identity preservation but not on the other metrics, since PyGAN tends to keep the results close to the input photo (and thus cannot perform large age changes).  

To measure identity preservation, we show the input and output photos and ask if the two contain the same person.
To measure age accuracy, we show the output photo and ask the age of the person, selected from a list of age ranges. To measure overall quality, we show an input photo, and below it a PyGAN result and our result side-by-side in a randomized order, and ask which result is a better version of the input person in the target age range. We used Amazon Mechanical Turk to collect answers for 20 randomly selected images from FFHQ-Aging dataset, repeating each question 5 times, for a total of 500 unique answers. We show the user study interface in supplemental material.

\begin{table}[t]
\centering
\setlength{\tabcolsep}{12pt}
\begin{tabular}{lcc}
\toprule
Age range:     & \multicolumn{2}{c}{50--69}                    \\
\cmidrule(lr){2-3}
               & PyGAN~\cite{yang2017learning} & Ours          \\
\midrule
Same identity  & \textbf{19}                   & 13            \\
Age difference & 23.1                          & \textbf{6.9}  \\
\addlinespace[0.3em]
Overall better & 4                             & \textbf{16}   \\
\bottomrule
\end{tabular}
\caption{User study results vs. PyGAN~\cite{yang2017learning}. PyGAN is expectedly better at identity preservation, at the cost of not generating the target age (mean age difference 23.1, compared to our 6.9). When asked which is better overall, users prefered our results in 16 out of 20 cases.}
\label{tbl:pygan}
\end{table}

\begin{table}[t]
\centering
\setlength{\tabcolsep}{3pt}
\begin{tabular}{lcccccccc}
\toprule
Age range:     & \multicolumn{2}{c}{15--19} & \multicolumn{2}{c}{30--39}  & \multicolumn{2}{c}{50--69} & \multicolumn{2}{c}{All}      \\
\cmidrule(lr){2-3} \cmidrule(lr){4-5} \cmidrule(lr){6-7} \cmidrule(lr){8-9}
               & IPCGAN & Ours              & IPCGAN      & Ours          & IPCGAN      & Ours         & IPCGAN       & Ours          \\
\midrule
Same identity  & 50     & 50                & \textbf{50} & 45            & \textbf{50} & 41           & \textbf{150} & 136           \\
Age difference & 19.3   & \textbf{12.7}     & 20.0        & \textbf{11.6} & 28.4        & \textbf{9.8} & 22.6         & \textbf{11.3} \\
\addlinespace[0.3em]
Overall better & 9      & \textbf{40}       & 8           & \textbf{42}   & 10          & \textbf{38}  & 27           & \textbf{120}  \\
\bottomrule
\end{tabular}
\caption{User study results vs. IPCGAN~\cite{wang2018face} for three age groups. IPCGAN is expectedly better at identity preservation, at the cost of not generating the target age (mean age difference 22.6, compared to our 11.3). When asked which is better overall, users prefered our results in 120 out of 150 cases.}
\label{tbl:ipcgan}
\end{table}

User study results are presented in \Cref{tbl:pygan}. 
As expected, PyGAN preserves subject identity more often (in 19 out of 20 cases, compared to 13 for our method). This comes at a cost of much larger age gaps: the perceived age of PyGAN results is on average 23.1 years away from the target age, compared to 6.9 years for our results.
Since identity preservation and age preservation may conflict, we also asked participants to evaluate which result is better overall.
For 16 out of 20 test photos, our results were rated as better than PyGAN.

In a second user study, we compare our results to those of IPCGAN trained on CACD dataset. We report results per age range, as well as overall results (\Cref{tbl:ipcgan}). %
We collected answers for 3 age ranges, 50 randomly selected images per range, repeating each question 3 times, for a total of 2250 unique answers. Similarly to PyGAN, IPCGAN better preserves identity (in 100\% of the cases) at the cost of age inaccuracies (results are on average 22.6 years away from the target age).
When asked which result is better overall, participants picked our results in 120 cases, compared to 27 for IPCGAN.

\begin{figure}[t!]
\centering
    \setlength\tabcolsep{0pt}
    \renewcommand{\arraystretch}{0.25}
    \begin{tabular}{>{\centering\arraybackslash}m{0.08\textwidth}
                    >{\centering\arraybackslash}m{0.45\textwidth}
                    >{\centering\arraybackslash}m{0.01\textwidth}
                    >{\centering\arraybackslash}m{0.45\textwidth}}
    \scalebox{0.6}{StarGAN} & \includegraphics[width=0.45\textwidth]{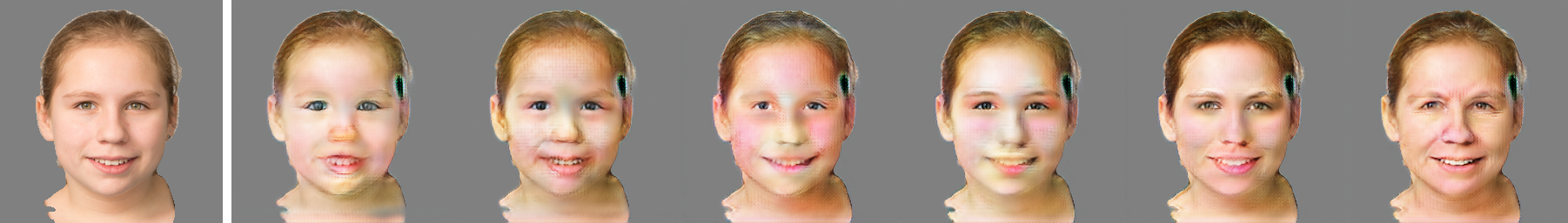} & & \includegraphics[width=0.45\textwidth]{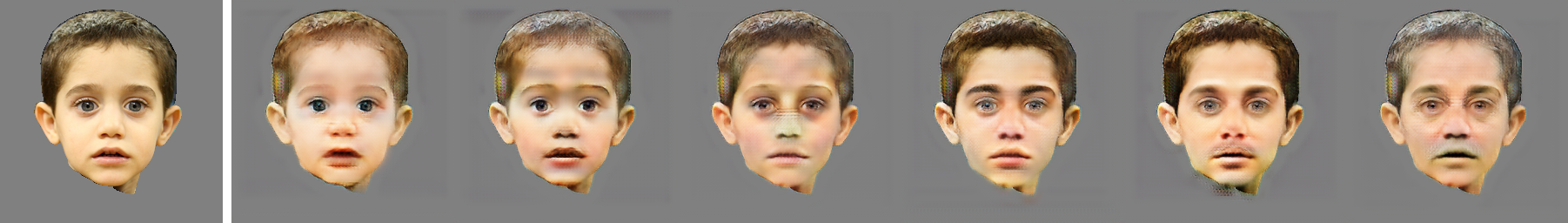}\tabularnewline
    \scalebox{0.6}{STGAN} & \includegraphics[width=0.45\textwidth]{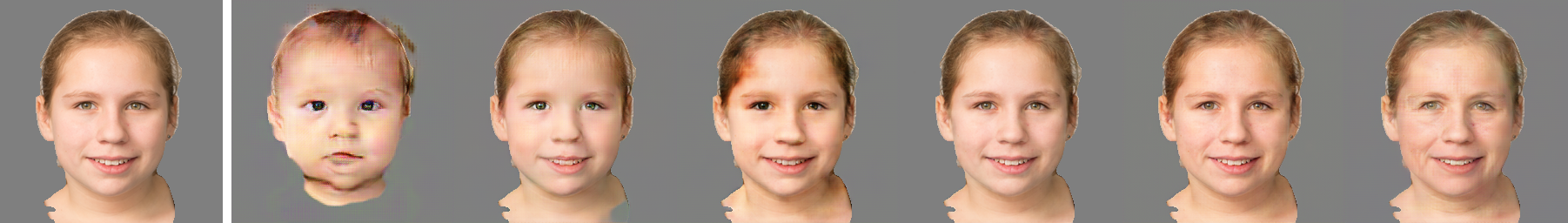} & & \includegraphics[width=0.45\textwidth]{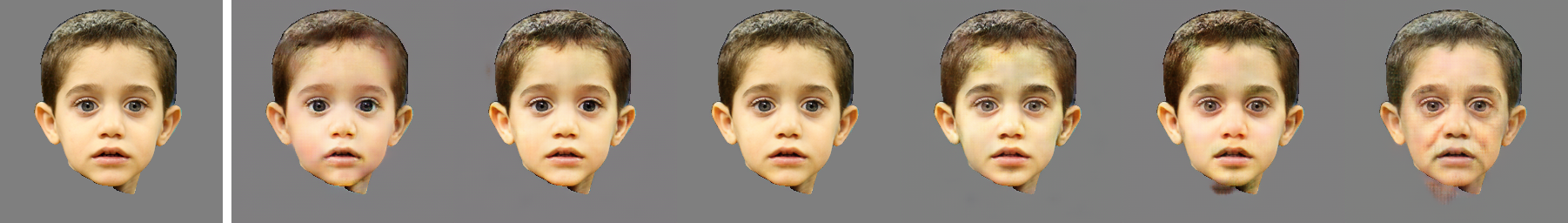}\tabularnewline
    \scalebox{0.6}{Ours} & \includegraphics[width=0.45\textwidth]{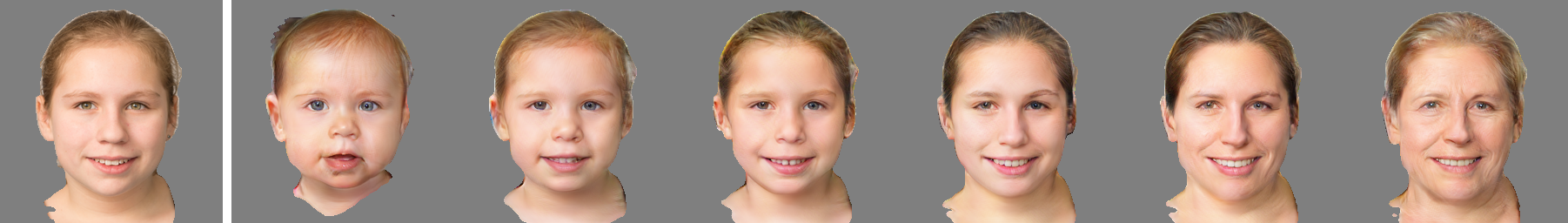} & & \includegraphics[width=0.45\textwidth]{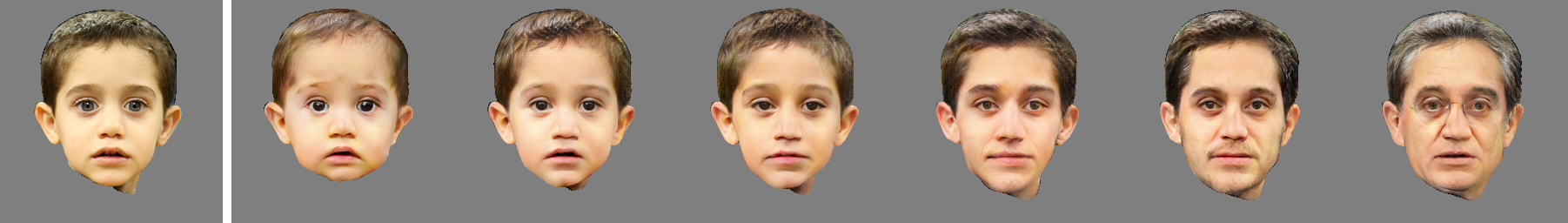}\tabularnewline
    \scalebox{0.6}{StarGAN} & \includegraphics[width=0.45\textwidth]{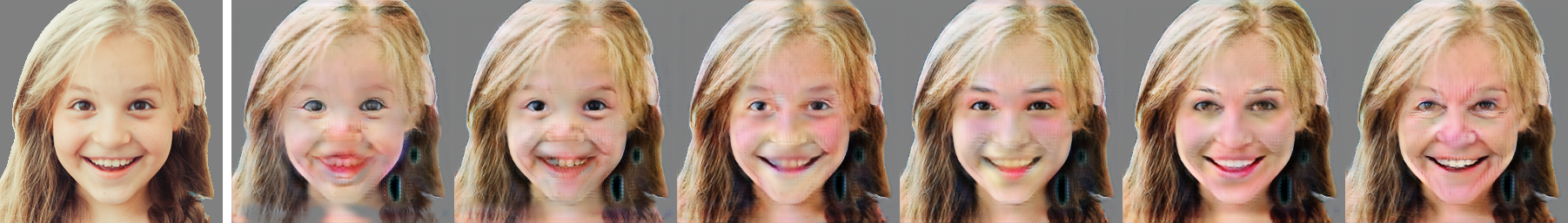} & & \includegraphics[width=0.45\textwidth]{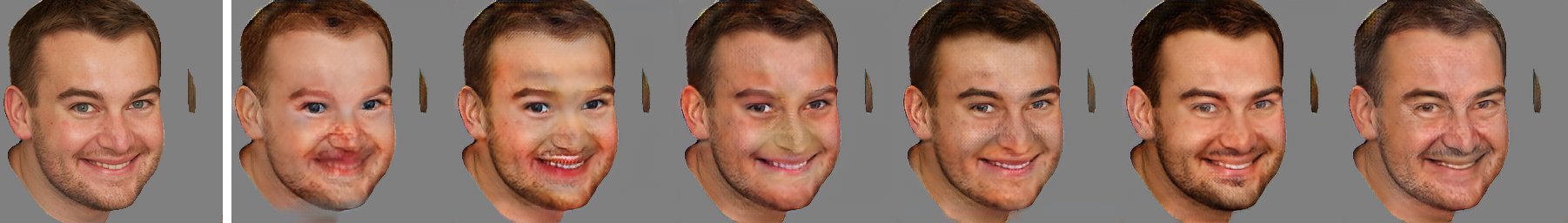}\tabularnewline
    \scalebox{0.6}{STGAN} & \includegraphics[width=0.45\textwidth]{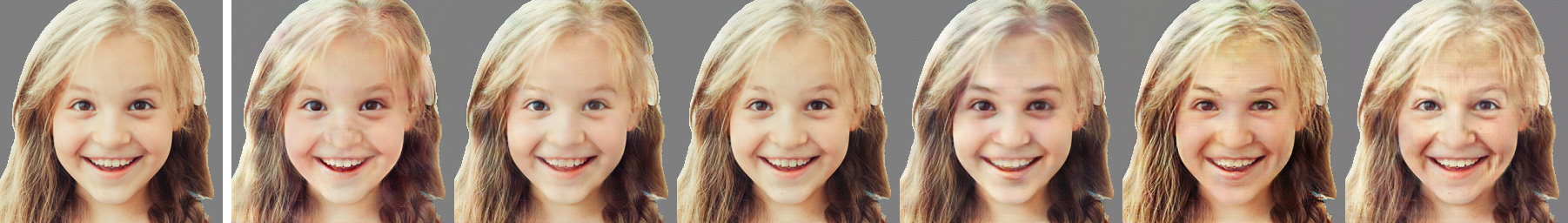} & & \includegraphics[width=0.45\textwidth]{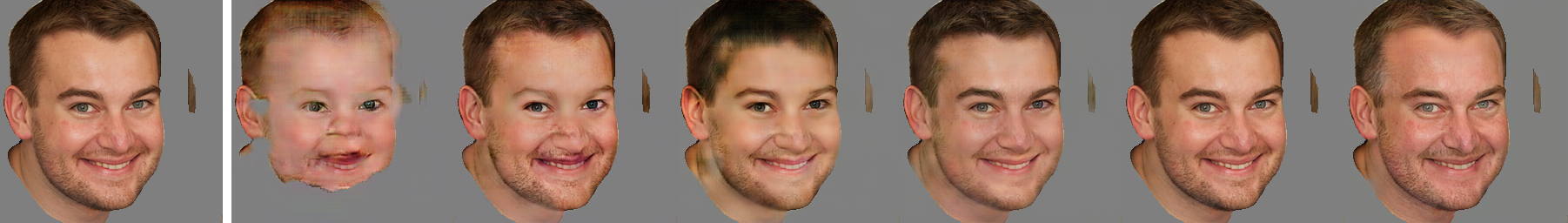}\tabularnewline
    \scalebox{0.6}{Ours} & \includegraphics[width=0.45\textwidth]{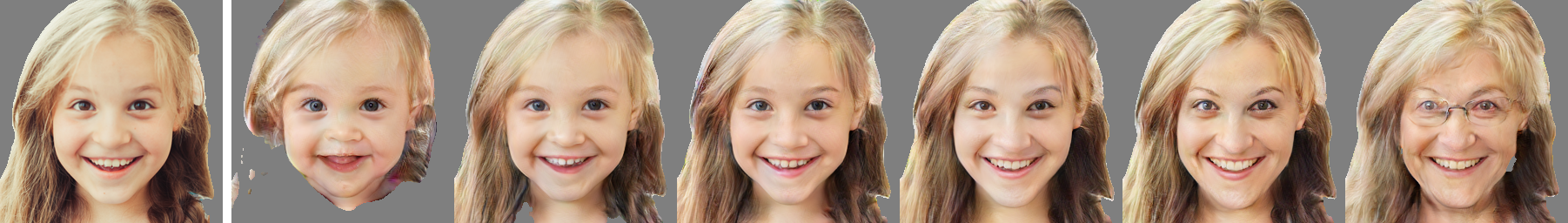} & & \includegraphics[width=0.45\textwidth]{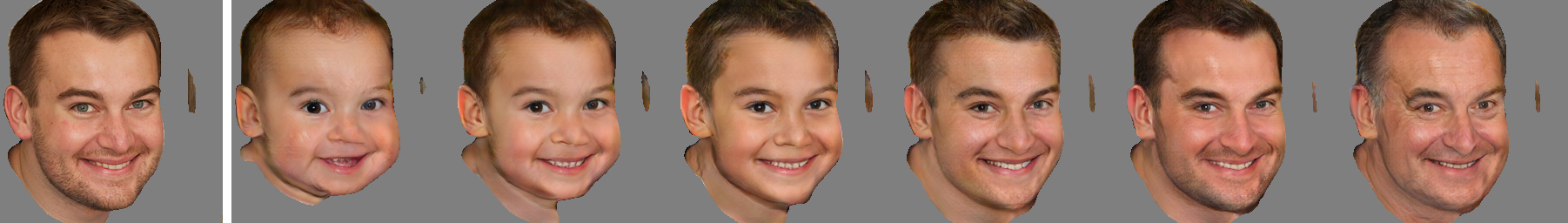}
    \end{tabular}
    \caption{Comparison with multi domain transfer methods. The leftmost column is the input, followed by transformations to age classes 0--2, 3--6, 7--9, 15--19, 30--39 and 50--69 respectively. Multi domain transfer methods struggle to model the gradual head deformation associated with age progression. Our method also produces better images in terms of quality and photorealism, while correctly modeling the growth of the head compared to StarGAN~\cite{choi2017stargan} and STGAN~\cite{liu2019stgan}.}
    \label{fig:multidomain}
\end{figure}

\textbf{Comparison with Multi-class Domain transfer methods}
To validate our claim that multi-domain transfer methods struggle with shape deformations, we compare our algorithm against 2 state-of-the-art baselines, StarGAN~\cite{choi2017stargan} and STGAN~\cite{liu2019stgan}. We retrain both algorithms on our FFHQ-Aging dataset using the same pre-processing procedure (see Sec.~\ref{alg_frame}) to mask background and clothes and the same sampling technique to compensate for dataset imbalances (see Sec.~\ref{alg_train}). \figurename~\ref{fig:multidomain} shows that although STGAN occasionally deforms the shape for the 0--2 class, both StarGAN and STGAN cannot produce a consistent shape transformation across age classes.

\begin{figure}[t]
\centering
    \setlength\tabcolsep{0pt}
    \renewcommand{\arraystretch}{0.25}
    \begin{tabular}{>{\centering\arraybackslash}m{0.0769\textwidth}
                    >{\centering\arraybackslash}m{0.0769\textwidth}
                    >{\centering\arraybackslash}m{0.0769\textwidth}
                    >{\centering\arraybackslash}m{0.0769\textwidth}
                    >{\centering\arraybackslash}m{0.0769\textwidth}
                    >{\centering\arraybackslash}m{0.0769\textwidth}
                    >{\centering\arraybackslash}m{0.0769\textwidth}
                    >{\centering\arraybackslash}m{0.0769\textwidth}
                    >{\centering\arraybackslash}m{0.0769\textwidth}
                    >{\centering\arraybackslash}m{0.0769\textwidth}
                    >{\centering\arraybackslash}m{0.0769\textwidth}
                    >{\centering\arraybackslash}m{0.0769\textwidth}
                    >{\centering\arraybackslash}m{0.0769\textwidth}}
    \cite{shen2019interpreting} \scalebox{0.5}{$\lambda \in [-3,3]$} &
    \multicolumn{12}{m{0.923\textwidth}}{\includegraphics[width=0.923\textwidth]{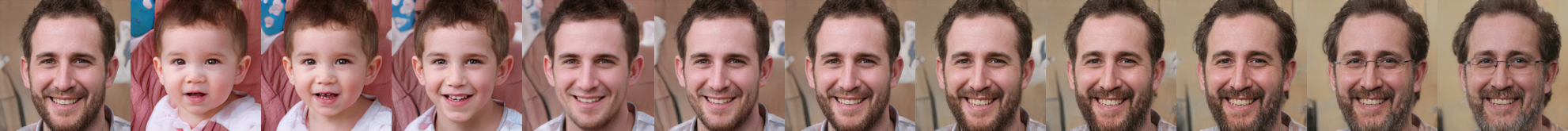}}\tabularnewline
    Ours &
    \multicolumn{12}{m{0.923\textwidth}}{\includegraphics[width=0.923\textwidth]{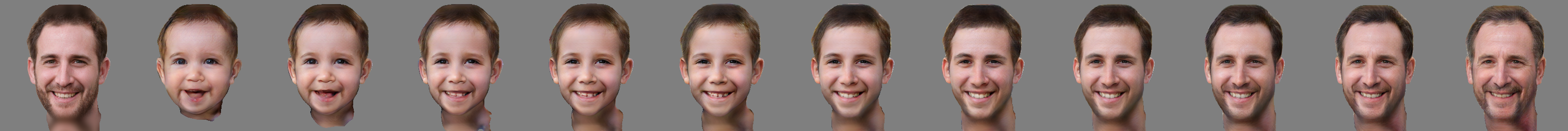}}\tabularnewline
    \cite{shen2019interpreting} \scalebox{0.5}{$\lambda \in [-3,3]$}&
    \multicolumn{12}{m{0.923\textwidth}}{\includegraphics[width=0.923\textwidth]{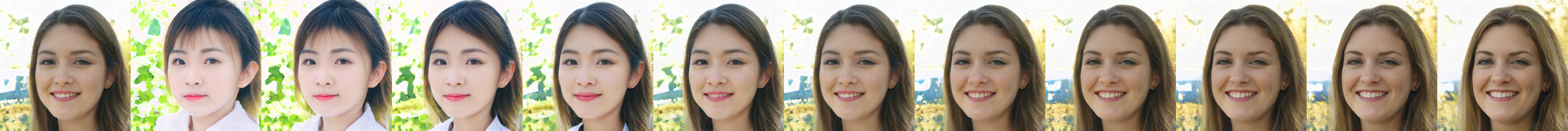}}\tabularnewline
    Ours &
    \multicolumn{12}{m{0.923\textwidth}}{\includegraphics[width=0.923\textwidth]{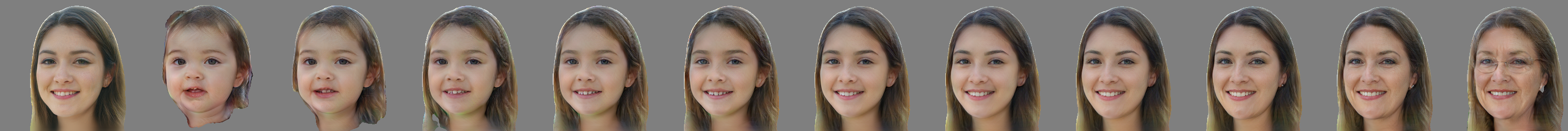}}\tabularnewline
    \cite{shen2019interpreting} \scalebox{0.5}{$ \lambda \in [-5,5]$}&
    \multicolumn{12}{m{0.923\textwidth}}{\includegraphics[width=0.923\textwidth]{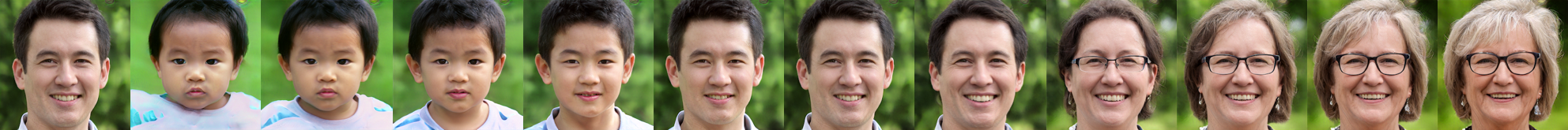}}\tabularnewline
    Ours &
    \multicolumn{12}{m{0.923\textwidth}}{\includegraphics[width=0.923\textwidth]{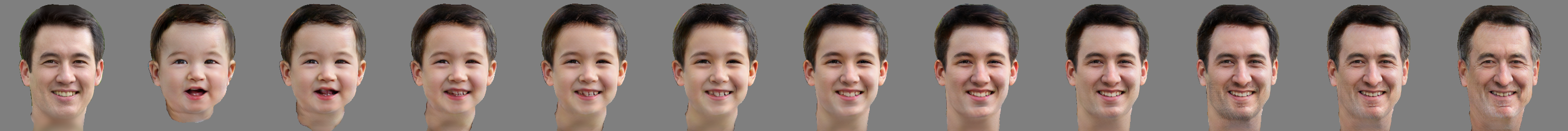}}\tabularnewline
    \scalebox{0.6}{\cite{zhu2019lia} \& \cite{shen2019interpreting}} \scalebox{0.5}{$\lambda \in [-20,20]$}&
    \multicolumn{12}{m{0.923\textwidth}}{\includegraphics[width=0.923\textwidth]{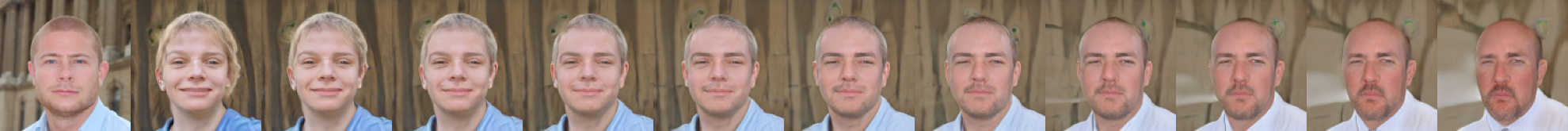}}\tabularnewline
    Ours &
    \multicolumn{12}{m{0.923\textwidth}}{\includegraphics[width=0.923\textwidth]{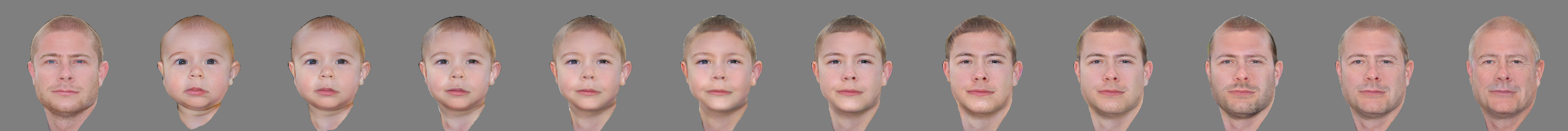}}\tabularnewline
    & \scriptsize{Input} & \scriptsize{0-2} & & \scriptsize{3-6} & & \scriptsize{7-9} & & \scriptsize{15-19} & & \scriptsize{30-39} & & \scriptsize{50-69}\tabularnewline
    \end{tabular}
    \caption{Comparison with InterFaceGAN~\cite{shen2019interpreting}. \textbf{Age cluster legend only applies to our method}. Rows 1,3,5: results on StyleGAN generated images. Row 7: result on a real image, embedded into the StyleGAN latent space using LIA~\cite{zhu2019lia}. Rows 2,4,6: Our result on StyleGAN generated images. Bottom row: Our result on a real image. Existing state-of-the-art interpolation methods cannot maintain the identity (rows 1,3,5,7), gender (row 5) and ethnicity (rows 3,5) of the input image. In addition, as seen in rows 1 \& 3, using the same $\lambda$ values on different photos produces different age ranges.}
    \label{fig:traversal}
\end{figure}

\textbf{Latent space interpolation}
We show our method's ability to generalize and produce continuous age transformations by interpolation in the $\mathcal{W}_{age}$ latent space. Interpolation between two neighboring age classes is done by generating two age latent codes $w_{age}^t, w_{age}^{t+1}$, where $t, t+1$ are adjacent age classes, and then generating the desired interpolated code $w_{age} = (1-\alpha) \cdot w_{age}^t + \alpha \cdot w_{age}^{t+1}$. The rest of the process is identical to Sec.~\ref{sec:algo}. We compare our results on two possible setups. In the first case, we demonstrate that the StyleGAN~\cite{karras2019style} latent space paths found by InterFaceGAN~\cite{shen2019interpreting} cannot maintain identity, gender and race. We sample a latent code in $z$ space of StyleGAN, which generates a realistic face image. We use the latent space boundary from InterFaceGAN, which can change age by preserving gender, and edit the sampled latent code to produce both younger and older versions of the face using $\lambda \in [-3,+3]$ in $z$ space. 
In the second setup, we compare against real images embedded into StyleGAN's $w$ latent space using the LIA~\cite{zhu2019lia} framework. We then change the age of the embedded face by traversing on the latent space across the age boundary learnt with InterFaceGAN on $w$ space (this boundary was not gender conditioned). We use $\lambda \in [-20,+20]$ in $w$ space to generate younger and older versions. 
In \figurename~\ref{fig:traversal} we can see that despite the excellent photorealism of InterFaceGAN on generated faces, the person's identity is lost, and in some occasions the gender is lost too. In addition, note how InterFaceGAN requires different $\lambda$ values for each input in order to achieve full lifespan transformation, as opposed to our consistent age outputs in the traversal paths.

\begin{figure}[t]
\centering
    \setlength\tabcolsep{0pt}
    \renewcommand{\arraystretch}{0.25}
    \begin{tabular}{>{\centering\arraybackslash}m{0.48\textwidth}
                    >{\centering\arraybackslash}m{0.02\textwidth}
                    >{\centering\arraybackslash}m{0.48\textwidth}}
    \includegraphics[width=0.48\textwidth]{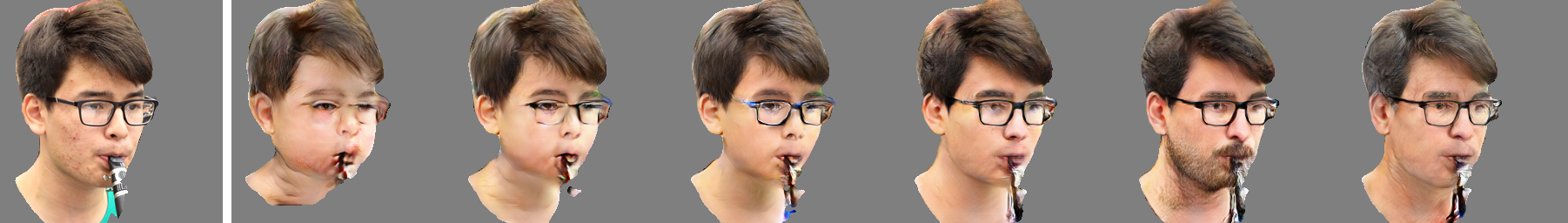} & & \includegraphics[width=0.48\textwidth]{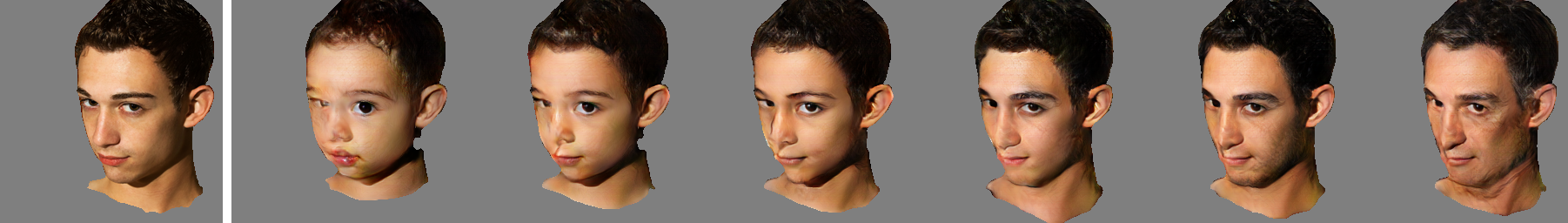}\tabularnewline
    \includegraphics[width=0.48\textwidth]{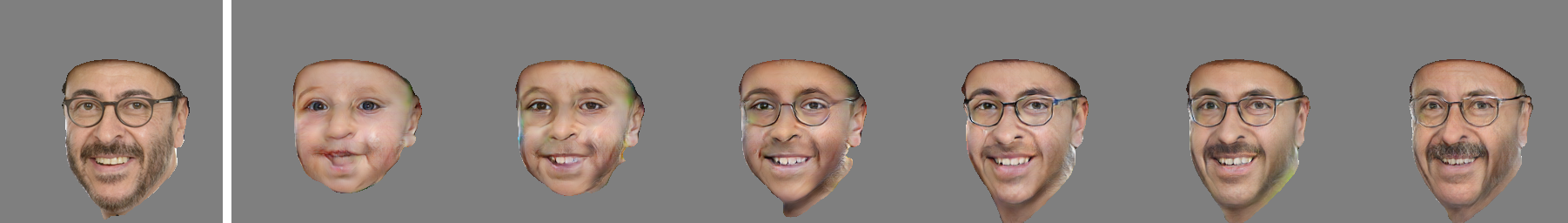} & & \includegraphics[width=0.48\textwidth]{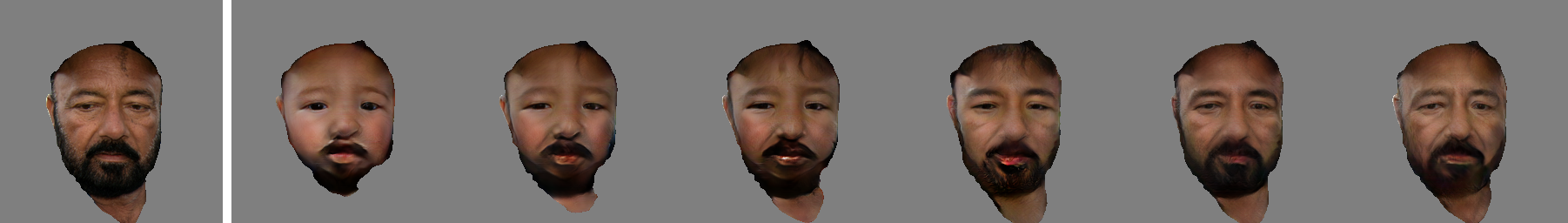}
    \end{tabular}
    \caption{Limitations. Our network struggles to generalize extreme poses (top row), removing glasses (left column), removing thick beards (bottom right) and occlusions (top left).}
    \label{fig:limitations}
\end{figure}

\subsection{Limitations}
While our network can generalize age transformations, it has limitations in generalizing for other potential cases such as extreme poses, removing glasses and thick beards when rejuvenating a person, and handling occluded faces. \figurename~\ref{fig:limitations} shows a representative example for each such case. We suspect that these issues stem from a combination of using just two downsampling layers in the identity encoder and the latent identity loss. The former creates relatively local feature maps, while the latter enforces the latent identity spatial representations of two difference age classes to be the same, which in turn, limits the network's ability to generalize for these cases. 

\section{Conclusions}
\label{sec:conc}
We presented an algorithm that produces reliable age transformations for ages 0--70. Unlike previous approaches, our framework learns to change both shape and texture as part of the aging process. The proposed architecture and training scheme accurately generalize age, and thus, we can produce results for ages never seen during training via latent space interpolation. In addition, we also introduced a new facial dataset that can be used by the vision community for various tasks. As demonstrated in our experiments, our method produces state-of-the-art results.
\ifanonymous

\else
    \section*{Acknowledgements}
    We wish to thank Xuan Luo and Aaron Wetzler for their valuable discussions and advice, and Thevina Dokka for her help in building the FFHQ-Aging dataset. This work was supported in part by Futurewei Technologies. O.F. was supported by the Brown Institute for Media Innovation.
    All images in this manuscript are licensed under creative commons license and were taken from the FFHQ dataset.
\fi

\bibliographystyle{splncs04}
\bibliography{ref.bib}

\ifarxiv
    \appendix
    \clearpage
\section*{Appendix}
In this appendix, we present our ethics and bias statement(\cref{sec:ethics}), additional details about the system architecture (\Cref{sec:net}), and the FFHQ-Aging dataset (\Cref{sec:dataset}). These are followed by additional results that show continuous age transformations (\Cref{sec:continuous}), ablation studies (\Cref{sec:abla}), evaluations of the framework generalization ability (\Cref{sec:gen})
and an extensive user study comparing our method to STGAN~\cite{liu2019stgan} and IPCGAN~\cite{wang2018face} on all of the trained 6 age clusters (\Cref{sec:user}).

\section{Ethics and Bias Statement}
\label{sec:ethics}
\subsection{Intended Use}
\begin{itemize}
    \item This algorithm is designed to hallucinate the aging process and produce an approximation of a person's appearance throughout his/her/their lifespan.
    \item The main use cases of this method are for art and entertainment purposes (CGI effects, Camera filters, etc.). This method might also be useful for more critical applications, e.g. approximating the appearance of missing people. However, we would like to stress that as a non perfect data-driven method, results might be inaccurate and biased. The output of our method should be critically analyzed by a trained professional, and not be treated as an absolute ground truth.
    \item \textbf{The results of this method should not be used as grounds for detention/arrest of a person or as any other form of legal evidence under any circumstances.}
\end{itemize}

\subsection{Algorithm and Data Bias}
We have devoted considerable efforts in our algorithm design to preserve the identity of the person in the input image, and to minimize the influence of the inherent dataset biases on the results. These measures include:
\begin{enumerate}
    \item Designing the identity encoder architecture to preserve the local structures of the input image.
    \item Including training losses that were designed to maintain the person's identity. 
    \begin{itemize}
        \item Latent Identity loss: encourages identity features that are consistent across ages. 
        \item Cycle loss: drives the network to reproduce the original image from any aged output.
        \item Self-reconstruction loss: makes the network learn to reconstruct the input when the target age class is the same as the source age class.
    \end{itemize}
    \item The FFHQ dataset contains gender imbalance within age classes. To prevent introducing these biases in the output, e.g. producing male facial features for females or vice versa, we have trained two separate models, one for males and one for females. The decision of which model to apply is left for the user. We acknowledge that this design choice restricts our algorithm from simulating the aging process of people whose gender is non-binary. Further work is required to make sure future algorithms will be able to simulate aging for the entire gender spectrum.
\end{enumerate}
Despite these measures, the network might still introduce other biases that we did not consider when designing the algorithm. If you spot any bias in the results, please reach out to help future research!

\section{Networks Architecture}
\label{sec:net}
Our framework consists of a generator, which contains the identity encoder, mapping network and the decoder, an age encoder and a discriminator. We describe the architecture of each component below.

\textbf{Identity encoder.} The identity encoder contains a $7 \times 7$ convolution layer that processes the input image. That layer is followed by two $3 \times 3$ 2-strided convolution layers that downsample the feature maps and four residual blocks~\cite{he2016deep} that produce the final identity features. Each convolution layer is followed by Pixel-norm~\cite{karras2017progressive}, which we empirically found to produce less artifacts than Instance-norm, and ReLU activation. We applied equalized learning rate~\cite{karras2017progressive} for each convolution layer. \Cref{tbl:id_encoder_arch} shows the Identity encoder architecture.

\textbf{Mapping network.} The mapping network is an 8 layer MLP network. It takes a $50 \times n$ input age code vector, where $n$ is the number of age classes, and outputs a 256 element age latent code. The input is first normalized with Pixel-norm~\cite{karras2017progressive}. Each fully connected layer is followed by a Leaky-ReLU activation and Pixel-norm. We omit the Leaky-ReLU activation for the last layer. We applied equalized learning rate~\cite{karras2017progressive} for each fully connected layer. The mapping network architecture can be seen in \Cref{tbl:map_net_arch}.

\textbf{Decoder.} Our decoder contains six styled convolution blocks~\cite{karras2019style} where we use bilinear upsampling in the last two blocks to return to the original image resolution. To reduce droplet artifacts, we replace each $3 \times 3$ convolution + AdaIN~\cite{huang2017adain} combination with a modulated convolution block proposed in StyleGAN2~\cite{Karras2019stylegan2}, omitting the noise input. Each modulated convolution layer is followed by a Leaky-ReLU activation and Pixel-norm, which we found to further help in reducing the droplet artifacts. The last layer is a $1 \times 1$ convolution that maps the final features of each pixel to RGB values. Equalized learning rate is used in all convolution blocks. Details of the decoder architecture are summarized in \Cref{tbl:decoder_arch}. 

\textbf{Age encoder.} The age encoder has a $7 \times 7$ convolution that takes the input image. It is followed by four $3 \times 3$ 2-strided convolution layers that downsample the feature maps, and a $1 \times 1$ convolution, that produces a feature map with $50 \times n$ output channels. A global average pooling is then applied to generate the age code vector. Each convolution layer, except for the last one, has a Leaky-ReLU activation. We don't use normalization in the age encoder. Equalized learning rate~\cite{karras2017progressive} was applied to each convolution layer. The full age encoder architecture can be found in \Cref{tbl:age_encoder_arch}.

\begin{table}[t!]
\begin{minipage}[t]{0.52\textwidth}
\centering
\begin{adjustbox}{max width=\textwidth}
\begin{tabular}{lcccr}
  \toprule
  Layer & Stride & Act. & Norm & Output Shape \\
  \midrule
  Input & -- & -- & -- & $256 \times 256 \times 3$\\
  \midrule
  Conv. $7 \times 7$ & 1 & ReLU & Pixel & $256 \times 256 \times 64$\\
  Conv. $3 \times 3$ & 2 & ReLU & Pixel & $128 \times 128 \times 128$\\
  Conv. $3 \times 3$ & 2 & ReLU & Pixel & $64 \times 64 \times 256$\\
  \midrule
  Res. Block & 1 & ReLU & Pixel & $64 \times 64 \times 256$\\
  Res. Block & 1 & ReLU & Pixel & $64 \times 64 \times 256$\\
  Res. Block & 1 & ReLU & Pixel & $64 \times 64 \times 256$\\
  Res. Block & 1 & ReLU & Pixel & $64 \times 64 \times 256$\\
  \bottomrule
\end{tabular}
\end{adjustbox}
\caption{Identity encoder architecture.}
\label{tbl:id_encoder_arch}
\end{minipage}
\hfill
\begin{minipage}[t]{0.4\textwidth}
\centering
\begin{adjustbox}{max width=\textwidth}
\begin{tabular}{lccc}
  \toprule
  Layer & Act. & Norm & Output Shape \\
  \midrule
  Age code & -- & Pixel & $50 \times n$\\
  \midrule
  Linear & LReLU & Pixel & $256$\\
  Linear & LReLU & Pixel & $256$\\
  Linear & LReLU & Pixel & $256$\\
  Linear & LReLU & Pixel & $256$\\
  Linear & LReLU & Pixel & $256$\\
  Linear & LReLU & Pixel & $256$\\
  Linear & LReLU & Pixel & $256$\\
  Linear & -- & Pixel & $256$\\
  \bottomrule
\end{tabular}
\end{adjustbox}
\caption{Mapping network architecture.}
\label{tbl:map_net_arch}
\end{minipage} \\

\begin{minipage}[t!]{0.46\textwidth}
\centering
\begin{adjustbox}{max width=\textwidth}
\begin{tabular}{lccr}
  \toprule
  Layer & Act. & Norm & Output Shape \\
  \midrule
  Idenity Features & -- & -- & $64 \times 64 \times 256$\\
  \midrule
  Styled Conv. & LReLU & Pixel & $64 \times 64 \times 256$\\
  Styled Conv. & LReLU & Pixel & $64 \times 64 \times 256$\\
  Styled Conv. & LReLU & Pixel & $64 \times 64 \times 256$\\
  Styled Conv. & LReLU & Pixel & $64 \times 64 \times 256$\\
  \midrule
  Styled Conv. & LReLU & Pixel & $64 \times 64 \times 128$\\
  Upsamle & -- & -- & $128 \times 128 \times 128$\\
  \midrule
  Styled Conv. & LReLU & Pixel & $128 \times  128 \times 64$\\
  Upsamle & -- & -- & $256 \times 256 \times 64$\\
  \midrule
  Conv. $1 \times 1$ & Tanh & -- & $256 \times 256 \times 3$\\
  \bottomrule
\end{tabular}
\end{adjustbox}
\caption{Decoder architecture.}
\label{tbl:decoder_arch}
\centering
\begin{adjustbox}{max width=\textwidth}
\begin{tabular}{lccr}
  \toprule
  Layer & Stride & Act. & Output Shape \\
  \midrule
  Input & -- & -- & $256 \times 256 \times 3$\\
  \midrule
  Conv. $7 \times 7$ & 1 & LReLU & $256 \times 256 \times 64$\\
  Conv. $3 \times 3$ & 2 & LReLU & $128 \times 128 \times 128$\\
  Conv. $3 \times 3$ & 2 & LReLU & $64 \times 64 \times 256$\\
  Conv. $3 \times 3$ & 2 & LReLU & $32 \times 32 \times 512$\\
  Conv. $3 \times 3$ & 2 & LReLU & $16 \times 16 \times 1024$\\
  Conv. $1 \times 1$ & 1 & -- & $16 \times 16 \times (50 \times n)$\\
  Global Pooling  & -- & -- & $1 \times 1 \times (50 \times n)$\\ 
  \bottomrule
\end{tabular}
\end{adjustbox}
\caption{Age encoder architecture.}
\label{tbl:age_encoder_arch}
\end{minipage}
\hfill
\begin{minipage}[t]{0.4825\textwidth}
\centering
\begin{adjustbox}{max width=\textwidth}
\begin{tabular}{lccr}
  \toprule
  Layer & Act. & Norm & Output Shape \\
  \midrule
  Input & -- & -- & $256 \times 256 \times 3$\\
  Conv. $1 \times 1$ & LReLU & -- & $256 \times 256 \times 64$\\
  \midrule
  Conv. $3 \times 3$ & LReLU & -- & $256 \times 256 \times 64$\\
  Conv. $3 \times 3$ & LReLU & -- & $256 \times 256 \times 128$\\
  Downsample & -- & -- & $128 \times 128 \times 128$\\
  \midrule
  Conv. $3 \times 3$ & LReLU & -- & $128 \times 128 \times 128$\\
 Conv. $3 \times 3$ & LReLU & -- & $128 \times 128 \times 256$\\
  Downsample & -- & -- & $64 \times 64 \times 256$\\
  \midrule
   Conv. $3 \times 3$ & LReLU & -- & $64 \times 64 \times 256$\\
  Conv. $3 \times 3$ & LReLU & -- & $64 \times 64 \times 512$\\
  Downsample & -- & -- & $32 \times 32 \times 512$\\
  \midrule
  Conv. $3 \times 3$ & LReLU & -- & $32 \times 32 \times 512$\\
  Conv. $3 \times 3$ & LReLU & -- & $32 \times 32 \times 512$\\
  Downsample & -- & -- & $16 \times 16 \times 512$\\
  \midrule
  Conv. $3 \times 3$ & LReLU & -- & $16 \times 16 \times 512$\\
  Conv. $3 \times 3$ & LReLU & -- & $16 \times 16 \times 512$\\
  Downsample & -- & -- & $8 \times 8 \times 512$\\
  \midrule
  Conv. $3 \times 3$ & LReLU & -- & $8 \times 8 \times 512$\\
  Conv. $3 \times 3$ & LReLU & -- & $8 \times 8 \times 512$\\
  Downsample & -- & -- & $4 \times 4 \times 512$\\
  \midrule
  Minibatch Stdev. & -- & -- & $4 \times 4 \times 513$\\
  Conv. $3 \times 3$ & LReLU & -- & $4 \times 4 \times 512$\\
  Conv. $4 \times 4$ & LReLU & -- & $1 \times 1 \times n$\\
  \bottomrule
\end{tabular}
\end{adjustbox}
\caption{Discriminator architecture.}
\label{tbl:disc_arch}
\end{minipage}
\vspace{-1cm}
\end{table}

\textbf{Discriminator.} We use the StyleGAN discriminator~\cite{karras2019style} architecture with minibatch standard deviation~\cite{karras2017progressive}. The first layer is a $1 \times 1$ convolution layer that generates a 64 channel feature map for each input pixel. This is followed by twelve $3 \times 3$ convolution layers~\cite{karras2019style}, we downsample the feature map after every other $3 \times 3$ block (6 times overall). After that we apply minibatch discrimination followed by a $3 \times 3$ convolution block and $4 \times 4$ convolution block with $n$ output channels in order to discriminate multiple classes as suggested by Liu \emph{et al.}~\cite{liu2019funit}. Leaky ReLU activations and Equalized learning rate are used in all convolution layers. We do not use normalization in the discriminator. \Cref{tbl:disc_arch} shows the detailed discriminator architecture.

\begin{figure}[t]
\begin{subfigure}[b]{0.68\textwidth}
\centering
\includegraphics[width=\textwidth,trim={0 0.85cm 0 0.7cm},clip]{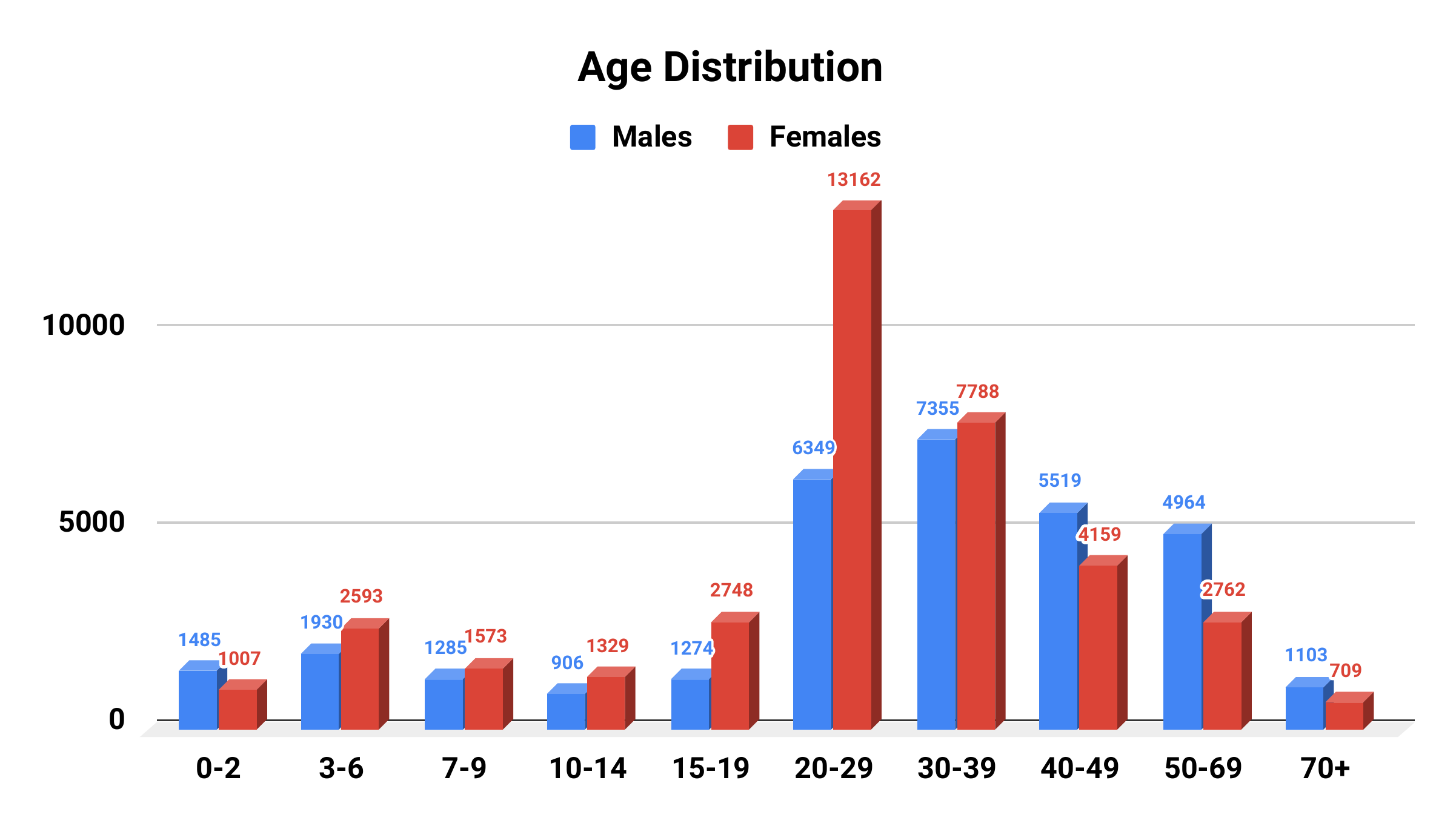}
\caption{}
\label{fig:hist}
\end{subfigure}
\hfill
\begin{subfigure}[b]{0.3\textwidth}
\centering
\begin{adjustbox}{max width=\textwidth}
\begin{tabular}{ccc}
  \toprule
  Age Class & Males & Females \\
  \midrule
  0--2 & 1237 & 804\\
  3--6 & 1631 & 2169\\
  7--9 & 1005 & 1234\\
  15--19 & 930 & 1957\\
  30--39 & 5512 & 5848\\
  50--69 & 3917 & 2054\\
  \bottomrule
\end{tabular}
\end{adjustbox}
\caption{}
\label{tbl:training_images}
\end{subfigure}
\caption{FFHQ-Aging dataset details. Left: age distributions for males and females for the raw dataset. Right: number of training images for each anchor age class after pruning. The majority of training classes contain more than 1,000 images, which we found sufficient for training our model.}
\end{figure}

\section{FFHQ-Aging Dataset Details}
\label{sec:dataset}

\figurename~\ref{fig:hist} shows the age distribution of images in the raw FFHQ-Aging dataset for males and females. \Cref{tbl:training_images} shows the number of training images for each age class after the data cleaning process described in Sec. 4 of the main paper.

To align the images, we use the same data alignment technique as Karras \emph{et al.}~\cite{karras2017progressive} (see \figurename \ 8e in their paper), which was also used to align the original FFHQ dataset. We mirror pad the image boundaries and then blur them. Then, we use the eyes and mouth landmark locations to select an oriented crop area according to
\begin{equation*}
    \begin{aligned}
    & x' = e_r - e_l \\
    & y' = \frac{1}{2}(m_r + m_l) - \frac{1}{2}(e_r + e_l) \\
    & c = \frac{1}{2}(e_r + e_l) - 0.1 \cdot y' \\
    & s = \max(4.0 \cdot |\text{Normalize}(x')|, 4.4 \cdot |\text{Normalize}(y')|) \\
    & x = \frac{s}{2} \cdot (\text{Normalize}(x' - \text{Rotate90}(y'))) \\
    & y = \text{Rotate90}(x) \\
    &\text{Box} = [c - x - y, c - x + y, c + x + y, c + x - y]
    \end{aligned}
\end{equation*}
Where $e_l, e_r$ are the landmarks for the left and right eyes respectively, $m_l, m_r$ are the landmarks for left and right corners of the mouth, "Normalize" is vector normalization, $s$ is the size of the box, and $c$ is the center of the cropping box.
In order to make sure we obtain the full head that also includes the neck, we took slightly larger crops then the original FFHQ dataset, our scale factor for $y'$ is 4.4 as opposed to 3.6 which was used originally.

\section{Continuous Age Transformations}
\label{sec:continuous}
We generated continuous lifespan age transformations by interpolating 24 output images between each neighboring age class anchors. See videos in the project's website\footnote{\href{https://grail.cs.washington.edu/projects/lifespan_age_transformation_synthesis/}{Lifespan Age Transformation Synthesis Project Website}} and \Cref{fig:video_frames_69222,fig:video_frames_69235} for the results. 

\section{Ablation Studies}
\label{sec:abla}
We performed two ablation studies in order to prove our main claims. In the first study we show the importance of using multiple age classes as anchors in order to learn a latent space $\mathcal{W}_{age}$ that will allow for continuous age transformations. We trained two additional models, one with age classes 0--2 \& 50--69 as the only anchors and one with age classes 0--2, 15--19 \& 50--69 as the anchors. We then generated full lifespan transformation of 11 images from each model by interpolating missing anchor classes when needed along with interpolating one output image between each two base classes. \figurename~\ref{fig:class_ablation} shows how additional anchor classes are crucial in creating reliable and plausible lifespan age transformations.

\begin{figure}[t]
\centering
    \setlength\tabcolsep{0pt}
    \renewcommand{\arraystretch}{0.25}
    \begin{tabular}{>{\centering\arraybackslash}m{0.0769\textwidth}
                    >{\centering\arraybackslash}m{0.0769\textwidth}
                    >{\centering\arraybackslash}m{0.0769\textwidth}
                    >{\centering\arraybackslash}m{0.0769\textwidth}
                    >{\centering\arraybackslash}m{0.0769\textwidth}
                    >{\centering\arraybackslash}m{0.0769\textwidth}
                    >{\centering\arraybackslash}m{0.0769\textwidth}
                    >{\centering\arraybackslash}m{0.0769\textwidth}
                    >{\centering\arraybackslash}m{0.0769\textwidth}
                    >{\centering\arraybackslash}m{0.0769\textwidth}
                    >{\centering\arraybackslash}m{0.0769\textwidth}
                    >{\centering\arraybackslash}m{0.0769\textwidth}
                    >{\centering\arraybackslash}m{0.0769\textwidth}}
    \scalebox{0.6}{2 Classes} &
    \multicolumn{12}{m{0.923\textwidth}}{\includegraphics[width=0.923\textwidth]{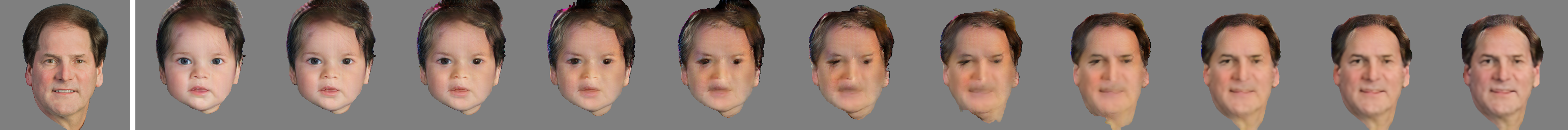}}\tabularnewline
    \scalebox{0.6}{3 Classes} &
    \multicolumn{12}{m{0.923\textwidth}}{\includegraphics[width=0.923\textwidth]{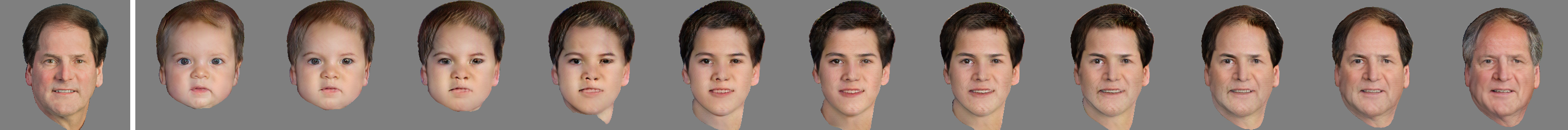}}\tabularnewline
    \scalebox{0.6}{6 Classes} \scalebox{0.6}{(ours)} &
    \multicolumn{12}{m{0.923\textwidth}}{\includegraphics[width=0.923\textwidth]{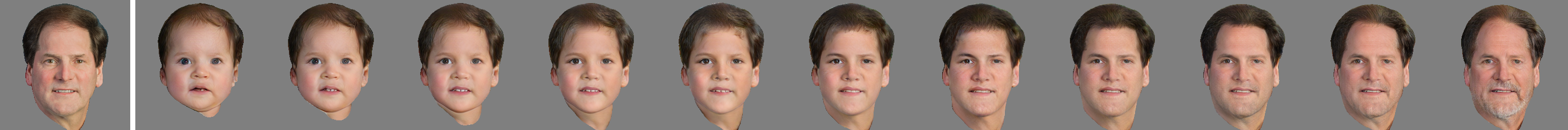}}\tabularnewline
    & \scriptsize{Input} & \scriptsize{0-2} & & \scriptsize{3-6} & & \scriptsize{7-9} & & \scriptsize{15-19} & & \scriptsize{30-39} & & \scriptsize{50-69}\tabularnewline
    \end{tabular}
    \caption{Anchor classes ablation study. We show latent interpolation on models trained on 2 anchor classes (top row), 3 anchor classes (middle row) and 6 anchor classes (bottom row). Increasing the number of anchor classes greatly improves the framework's ability to generate high quality age transformations over full lifespan.
    }
    \label{fig:class_ablation}
\end{figure}

In the second study, we examined importance of our design choices in constructing the input age vector code space $\mathcal{Z}$. We show the connection between the structure of $\mathcal{Z}$ to the ability of the age latent space $\mathcal{W}_{age}$ to span all possible ages. Specifically, we show the importance of using multiple vector elements to represent each age class as well as the importance of adding noise to the one-hot input signal. We trained two additional models on all 6 anchor classes, one with 50 elements per age class, but with no added noise, and one with a single element per age class and no added noise. In \figurename~\ref{fig:age_code_ablation} we can see that although the anchor classes are always well represented within the latent space, both number of elements per age class and added noise, are important parts to ensure the continuity of $\mathcal{W}_{age}$ and high image quality.

\begin{figure}[t]
\centering
    \setlength\tabcolsep{0pt}
    \renewcommand{\arraystretch}{0.25}
    \begin{tabular}{>{\centering\arraybackslash}m{0.0769\textwidth}
                    >{\centering\arraybackslash}m{0.0769\textwidth}
                    >{\centering\arraybackslash}m{0.0769\textwidth}
                    >{\centering\arraybackslash}m{0.0769\textwidth}
                    >{\centering\arraybackslash}m{0.0769\textwidth}
                    >{\centering\arraybackslash}m{0.0769\textwidth}
                    >{\centering\arraybackslash}m{0.0769\textwidth}
                    >{\centering\arraybackslash}m{0.0769\textwidth}
                    >{\centering\arraybackslash}m{0.0769\textwidth}
                    >{\centering\arraybackslash}m{0.0769\textwidth}
                    >{\centering\arraybackslash}m{0.0769\textwidth}
                    >{\centering\arraybackslash}m{0.0769\textwidth}
                    >{\centering\arraybackslash}m{0.0769\textwidth}}
    \scalebox{0.5}{1 Element} \scalebox{0.5}{No Noise} &
    \multicolumn{12}{m{0.923\textwidth}}{\includegraphics[width=0.923\textwidth]{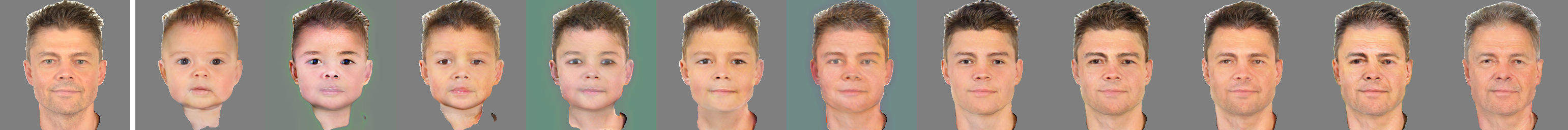}}\tabularnewline
    \scalebox{0.5}{50 Elements} \scalebox{0.5}{No Noise} &
    \multicolumn{12}{m{0.923\textwidth}}{\includegraphics[width=0.923\textwidth]{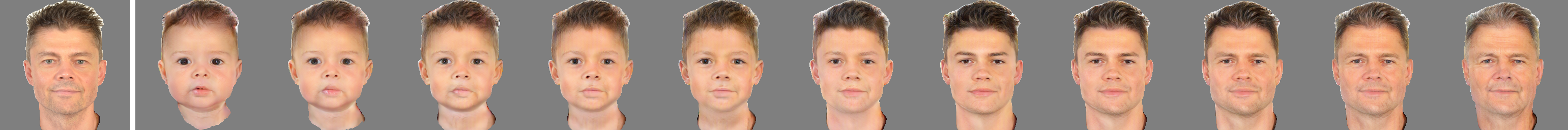}}\tabularnewline
    \scalebox{0.5}{50 Elements} \scalebox{0.5}{With Noise} &
    \multicolumn{12}{m{0.923\textwidth}}{\includegraphics[width=0.923\textwidth]{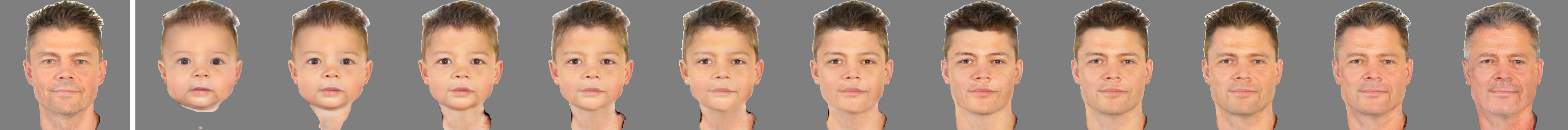}}\tabularnewline
    & \scriptsize{Input} & \scriptsize{0-2} & & \scriptsize{3-6} & & \scriptsize{7-9} & & \scriptsize{15-19} & & \scriptsize{30-39} & & \scriptsize{50-69}\tabularnewline
    \end{tabular}
    \caption{Age class representation ablation study. We show latent interpolation on models trained with one-hot representation with 1 element per age class (top row), one-hot representation with 50 elements per age class (middle row) and one-hot representation with 50 elements per age class and added gaussian noise (bottom row). Expanding the number of elements representing each age class allows representation of ages outside the anchor classes. Adding noise, further improves the image quality for interpolated outputs (Zoom in for details).}
    \label{fig:age_code_ablation}
\end{figure}

\subsection{Generalization Ability}
\label{sec:gen}

\begin{figure}[t!]
\centering
    \setlength\tabcolsep{0pt}
    \renewcommand{\arraystretch}{0.25}
    \begin{tabular}{>{\centering\arraybackslash}m{0.097\textwidth}
                    >{\centering\arraybackslash}m{0.00382\textwidth}
                   >{\centering\arraybackslash}m{0.097\textwidth}
                   >{\centering\arraybackslash}m{0.097\textwidth}
                    >{\centering\arraybackslash}m{0.097\textwidth}
                    >{\centering\arraybackslash}m{0.097\textwidth}
                    >{\centering\arraybackslash}m{0.02\textwidth}
                    >{\centering\arraybackslash}m{0.097\textwidth}
                    >{\centering\arraybackslash}m{0.00382\textwidth}
                    >{\centering\arraybackslash}m{0.097\textwidth}
                    >{\centering\arraybackslash}m{0.097\textwidth}
                    >{\centering\arraybackslash}m{0.097\textwidth}
                    >{\centering\arraybackslash}m{0.097\textwidth}}
    \multicolumn{3}{m{0.1978\textwidth}}{\includegraphics[width=0.1978\textwidth,trim={0 0 512px 0},clip]{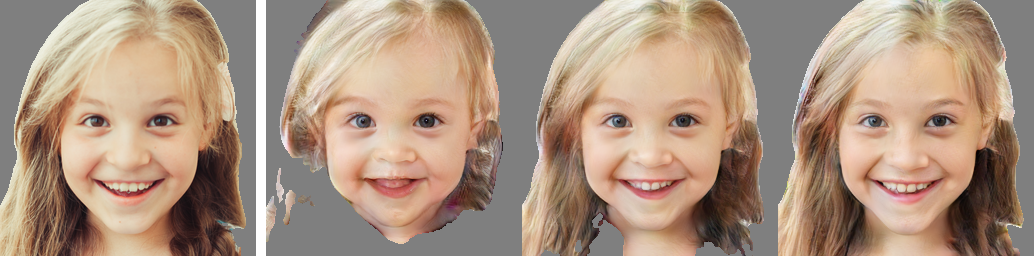}} & \includegraphics[width=0.097\textwidth]{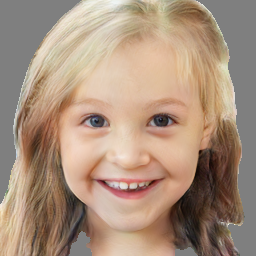} & \multicolumn{2}{m{0.194\textwidth}}{\includegraphics[width=0.194\textwidth,trim={522px 0 0 0},clip]{images/3_6_comp/69005_interp.png}} & & 
    
    \multicolumn{3}{m{0.1978\textwidth}}{\includegraphics[width=0.1978\textwidth,trim={0 0 512px 0},clip]{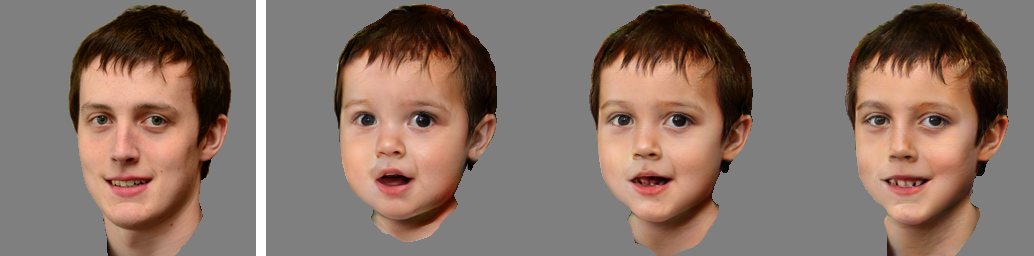}} & \includegraphics[width=0.097\textwidth]{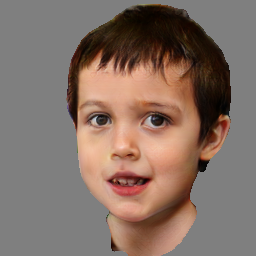} & \multicolumn{2}{m{0.194\textwidth}}{\includegraphics[width=0.194\textwidth,trim={522px 0 0 0},clip]{images/3_6_comp/69451_interp.png}}\tabularnewline
    
    \multicolumn{3}{m{0.1978\textwidth}}{\includegraphics[width=0.1978\textwidth,trim={0 0 512px 0},clip]{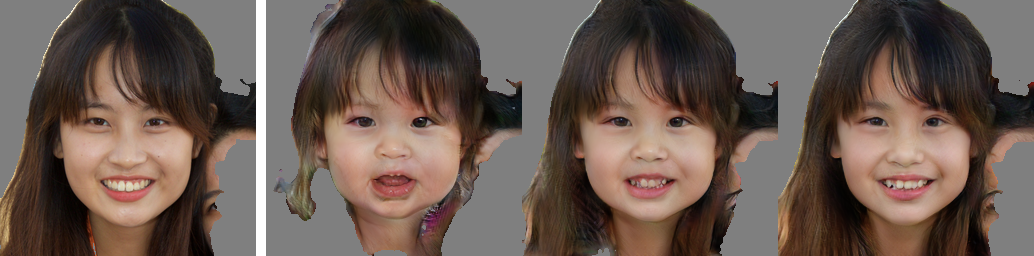}} & \includegraphics[width=0.097\textwidth]{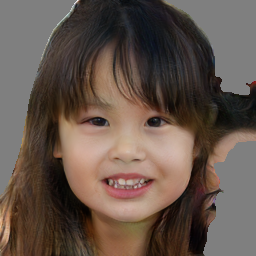} & \multicolumn{2}{m{0.194\textwidth}}{\includegraphics[width=0.194\textwidth,trim={522px 0 0 0},clip]{images/3_6_comp/69435_interp.png}} & &
    
    \multicolumn{3}{m{0.1978\textwidth}}{\includegraphics[width=0.1978\textwidth,trim={0 0 512px 0},clip]{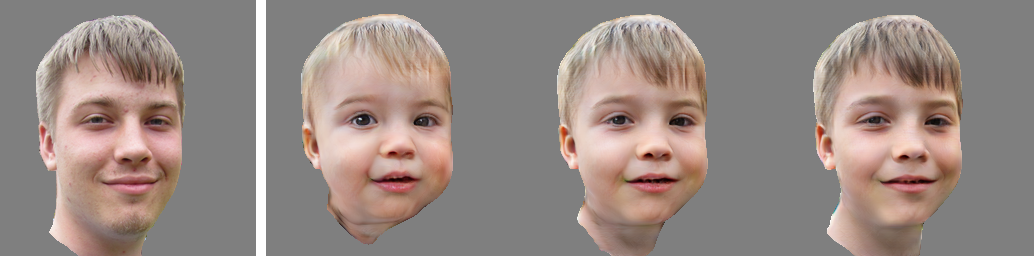}} & \includegraphics[width=0.097\textwidth]{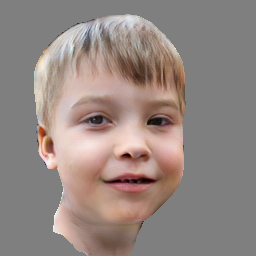} & \multicolumn{2}{m{0.194\textwidth}}{\includegraphics[width=0.194\textwidth,trim={522px 0 0 0},clip]{images/3_6_comp/69735_interp.png}}\tabularnewline
    
    \multicolumn{3}{m{0.1978\textwidth}}{\includegraphics[width=0.1978\textwidth,trim={0 0 512px 0},clip]{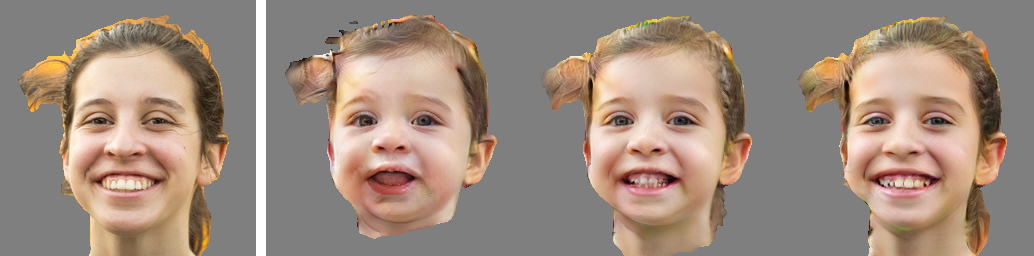}} & \includegraphics[width=0.097\textwidth]{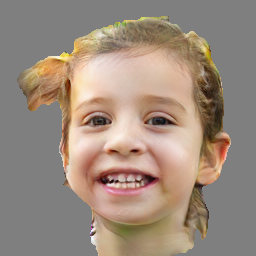} & \multicolumn{2}{m{0.194\textwidth}}{\includegraphics[width=0.194\textwidth,trim={522px 0 0 0},clip]{images/3_6_comp/69828_interp.png}} & &
    
    \multicolumn{3}{m{0.1978\textwidth}}{\includegraphics[width=0.1978\textwidth,trim={0 0 512px 0},clip]{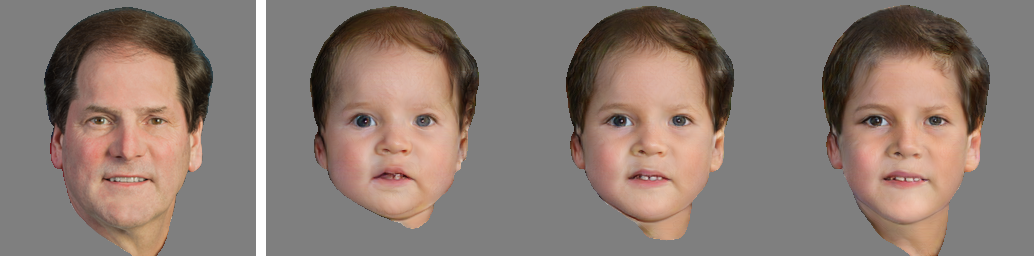}} & \includegraphics[width=0.097\textwidth]{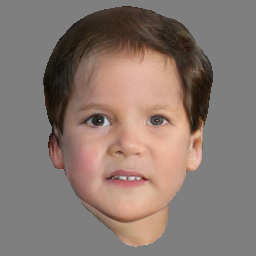} & \multicolumn{2}{m{0.194\textwidth}}{\includegraphics[width=0.194\textwidth,trim={522px 0 0 0},clip]{images/3_6_comp/69691_interp.png}}\tabularnewline
    
    \scriptsize{Input Image} & & \scriptsize{0--2} & \scriptsize{Trained 3--6} & \scriptsize{Interpolated 3--6} & \scriptsize{7--9} & & \scriptsize{Input Image} & & \scriptsize{0--2} & \scriptsize{Trained 3--6} & \scriptsize{Interpolated 3--6} & \scriptsize{7--9}
    \end{tabular}
    \caption{Linearity of age latent space. We compare the results of the network outputs for the 3--6 class vs. the network outputs for 3--6 class interpolated as the mid point between the 0--2 and 7--9 age latent vectors $0.5 \cdot w_{age}^{0-2} + 0.5 \cdot w_{age}^{7-9}$. The resemblance of the interpolated results to the trained results suggests that age is spanned quasi-linearly in the $\mathcal{W}_{age}$ latent space.}
    \label{fig:comp_3_6}
    \ifarxiv
    \vspace{-0.25cm}
    \fi
\end{figure}

To test our framework ability to generalize, we carried out two experiments. In the first experiment, we tested the generalization ability of the age latent space $\mathcal{W}_{age}$. We produced outputs for the 3--6 age class by interpolating it as the mid point of 0--2 and 7--9 age classes. We fed the decoder a latent age vector $\Tilde{w}_{age}^{3-6} = 0.5 \cdot w_{age}^{0-2} + 0.5 \cdot w_{age}^{7-9}$ and compare the results with the outputs for the trained 3--6 class. As can be seen in \figurename~\ref{fig:comp_3_6}, the similarity between the trained results and the interpolated results suggests that the learned age latent space, $\mathcal{W}_{age}$, is approximately linear w.r.t the target age input which contributes to the ability of the framework to generate results outside of the trained age classes. 

In the second experiment, we tested the generalization ability of the identity feature space. We feed the network images from the remaining 4 untrained classes in FFHQ-Aging, 10--14, 20--29, 40--49 \& 70+. \figurename~\ref{fig:untrained_classes} demonstrates our method's ability to produce high-quality results for unseen face structures from unseen age classes.

\begin{figure}[t!]
\centering
    \setlength\tabcolsep{0pt}
    \renewcommand{\arraystretch}{0.25}
    \begin{tabular}{>{\centering\arraybackslash}m{0.09\textwidth}
                    >{\centering\arraybackslash}m{0.129\textwidth}
                   >{\centering\arraybackslash}m{0.005\textwidth}
                   >{\centering\arraybackslash}m{0.129\textwidth}
                   >{\centering\arraybackslash}m{0.129\textwidth}
                    >{\centering\arraybackslash}m{0.129\textwidth}
                   >{\centering\arraybackslash}m{0.129\textwidth}
                   >{\centering\arraybackslash}m{0.129\textwidth}
                   >{\centering\arraybackslash}m{0.129\textwidth}}
    \scalebox{0.7}{10--14} &
    \multicolumn{8}{m{0.91\textwidth}}{\includegraphics[width=0.91\textwidth]{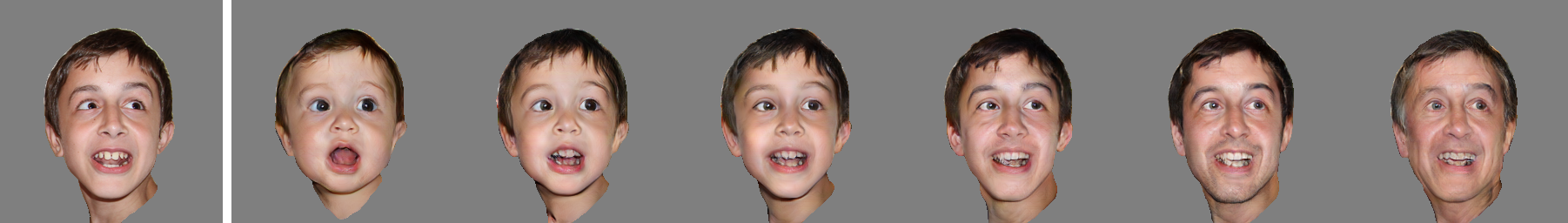}}\tabularnewline
    
    \scalebox{0.7}{10--14} &
    \multicolumn{8}{m{0.91\textwidth}}{\includegraphics[width=0.91\textwidth]{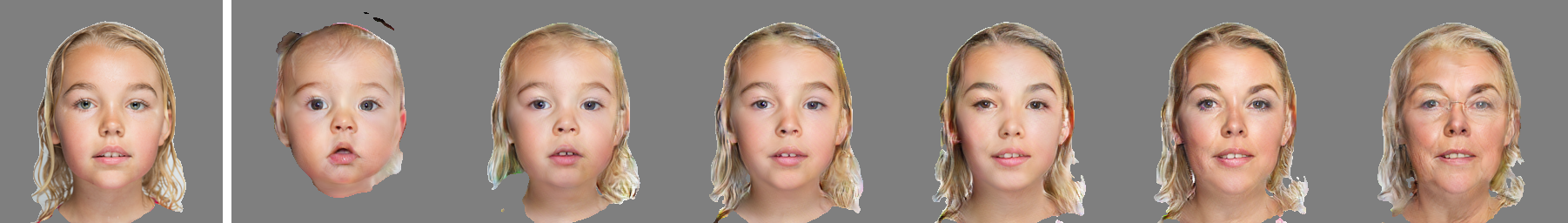}}\tabularnewline
    
    \scalebox{0.7}{20--29} &
    \multicolumn{8}{m{0.91\textwidth}}{\includegraphics[width=0.91\textwidth]{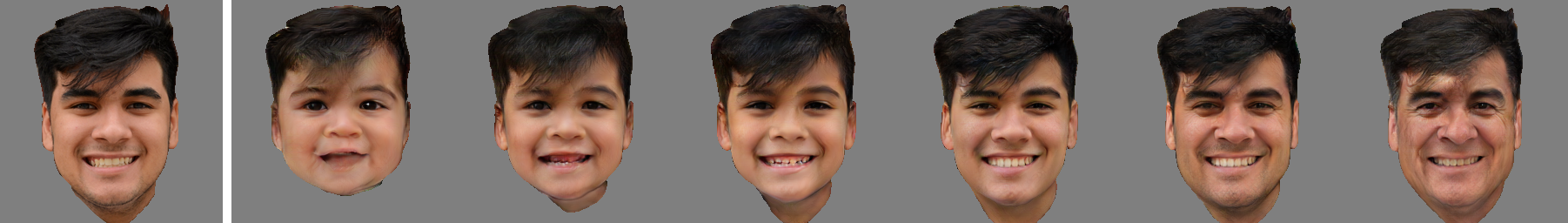}}\tabularnewline
    
    \scalebox{0.7}{20--29} &
    \multicolumn{8}{m{0.91\textwidth}}{\includegraphics[width=0.91\textwidth]{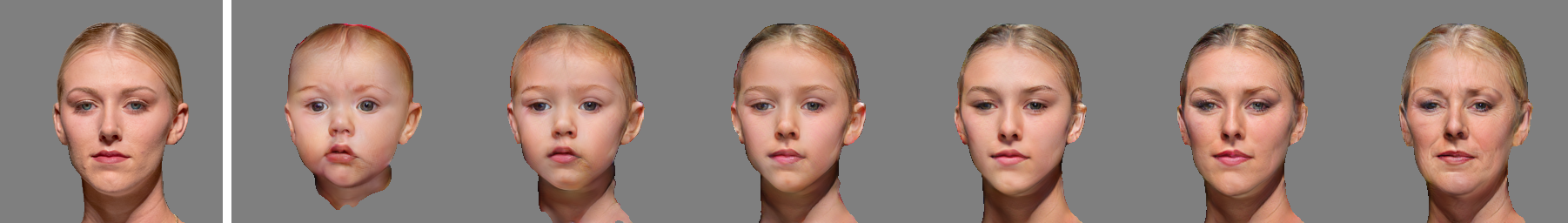}}\tabularnewline
    
    \scalebox{0.7}{40--49} &
    \multicolumn{8}{m{0.91\textwidth}}{\includegraphics[width=0.91\textwidth]{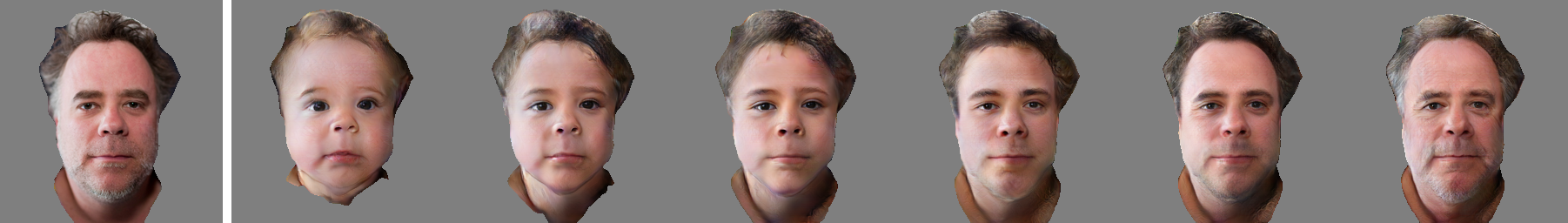}}\tabularnewline
    
    \scalebox{0.7}{40--49} &
    \multicolumn{8}{m{0.91\textwidth}}{\includegraphics[width=0.91\textwidth]{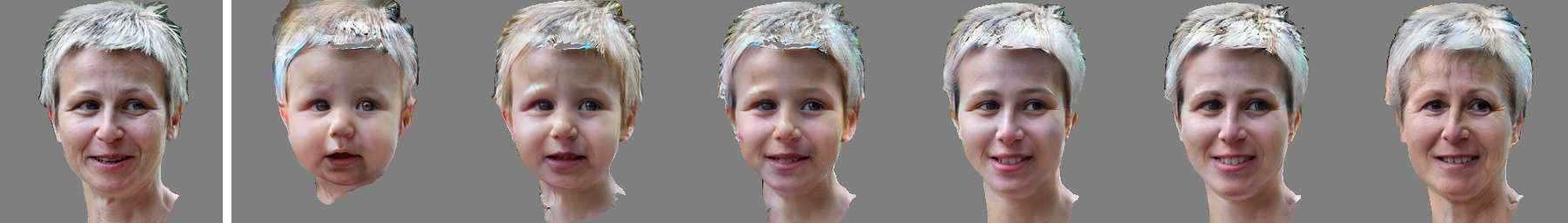}}\tabularnewline
    
    \scalebox{0.7}{70+} &
    \multicolumn{8}{m{0.91\textwidth}}{\includegraphics[width=0.91\textwidth]{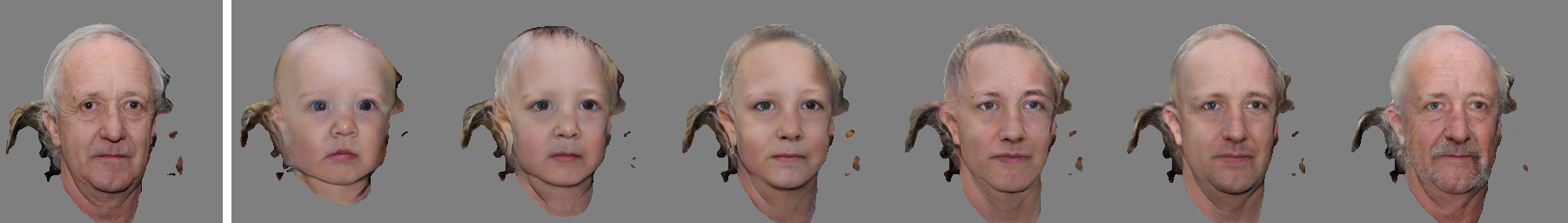}}\tabularnewline 
    
    \scalebox{0.7}{70+} &
    \multicolumn{8}{m{0.91\textwidth}}{\includegraphics[width=0.91\textwidth]{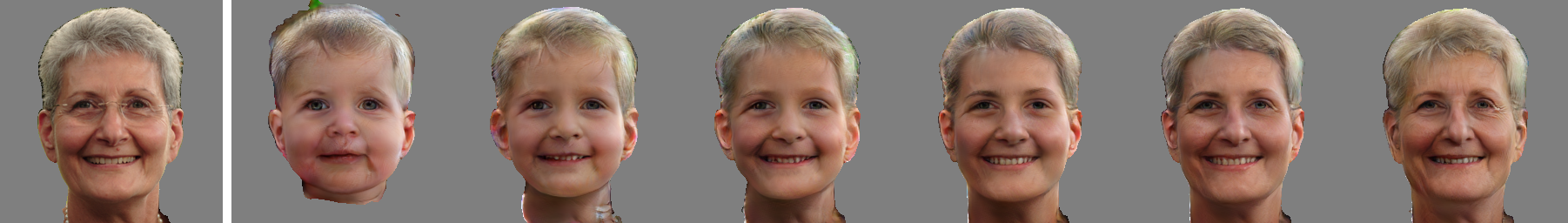}}\tabularnewline
    
    \scriptsize{Input Age} & \scriptsize{Input Image} & & \scriptsize{0-2} & \scriptsize{3-6} & \scriptsize{7-9} & \scriptsize{15-19} & \scriptsize{30-39} & \scriptsize{50-69}\tabularnewline
    \end{tabular}
    \caption{Results on inputs from untrained age classes. Note that masking artifacts are a result of the segmentation process, and were not caused due to our method.}
    \label{fig:untrained_classes}
    \ifarxiv
    \vspace{-0.25cm}
    \fi
\end{figure}

\subsection{User Studies}
\label{sec:user}

\begin{figure}[h!]
    \centering
    \includegraphics[width=0.325\textwidth,frame=0.1pt]{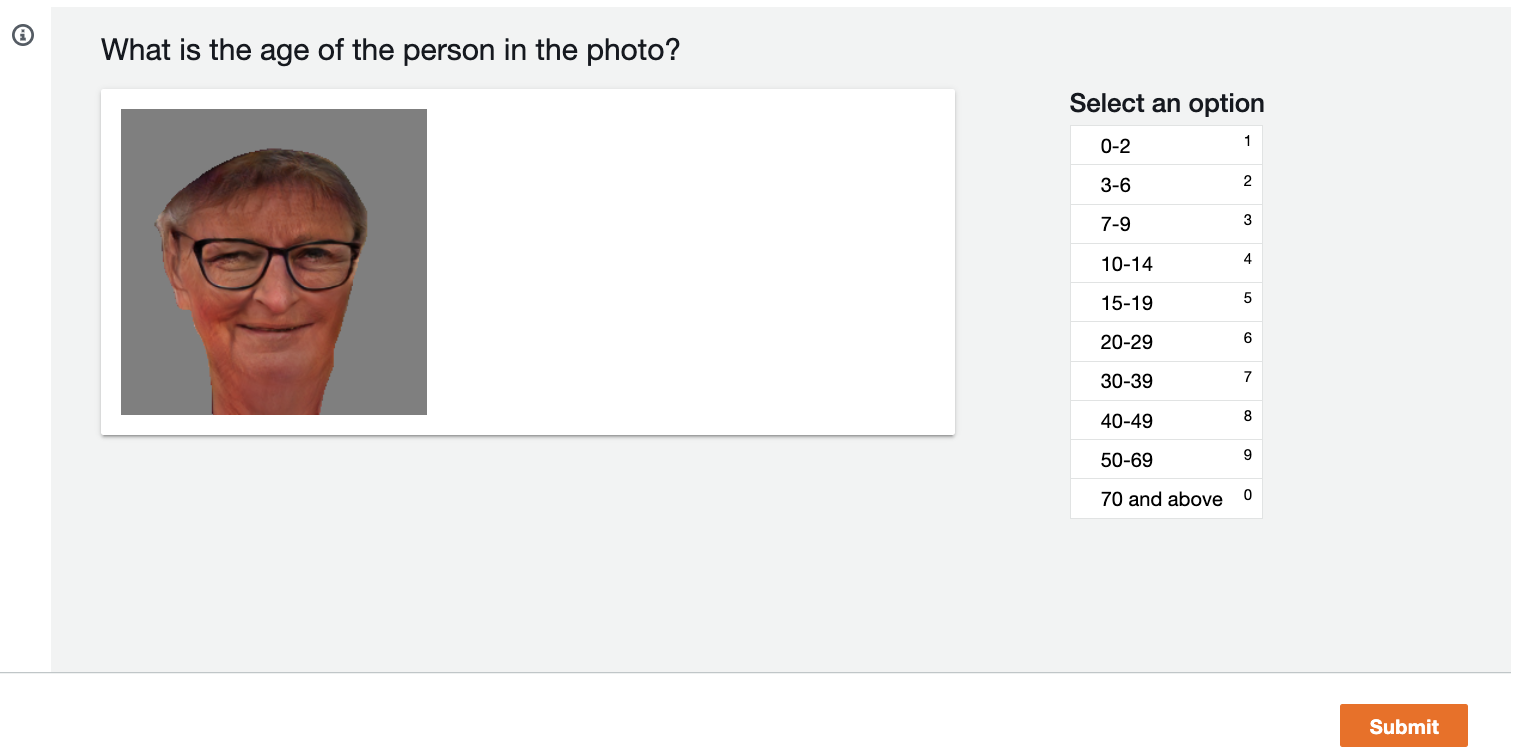}
    \includegraphics[width=0.325\textwidth,frame=0.1pt]{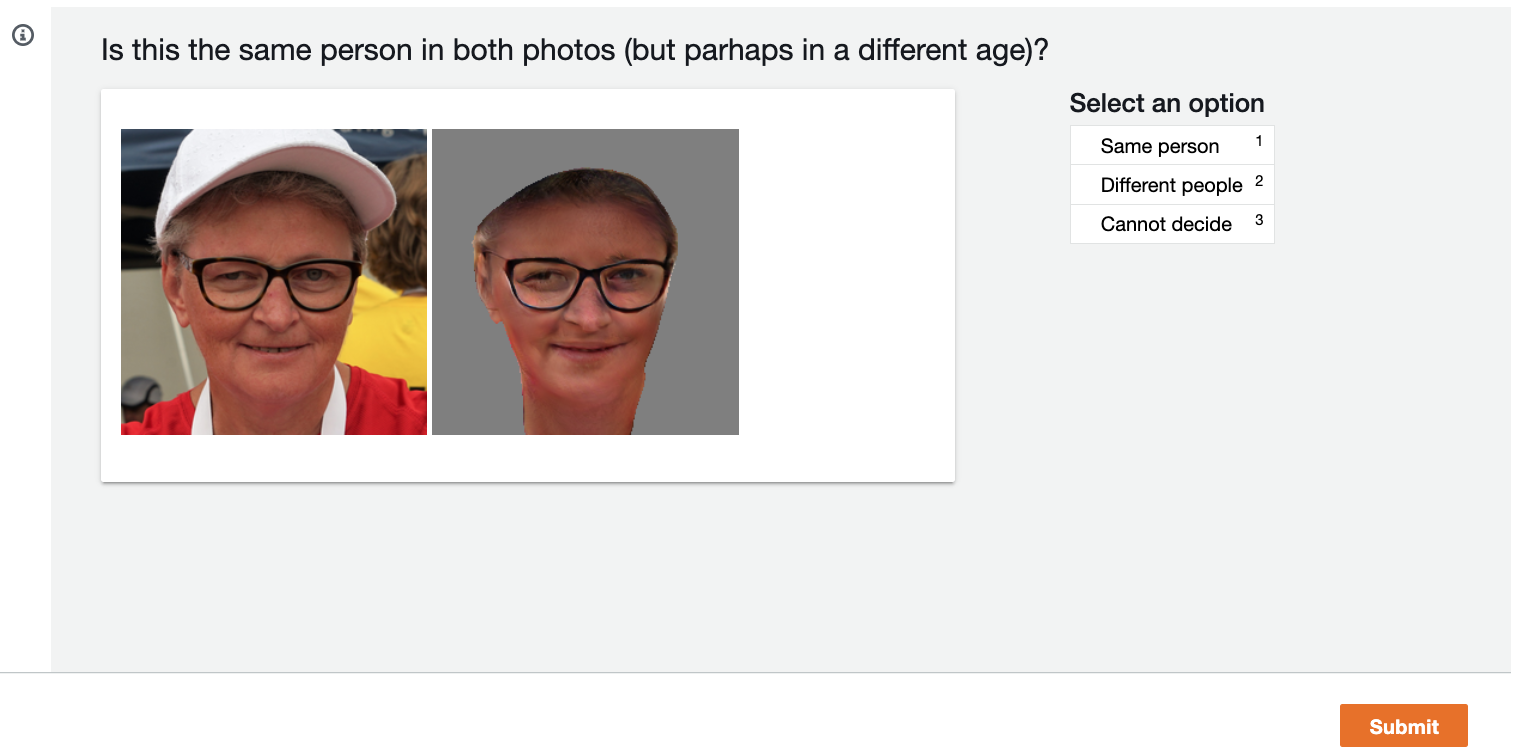}
    \includegraphics[width=0.325\textwidth,frame=0.1pt]{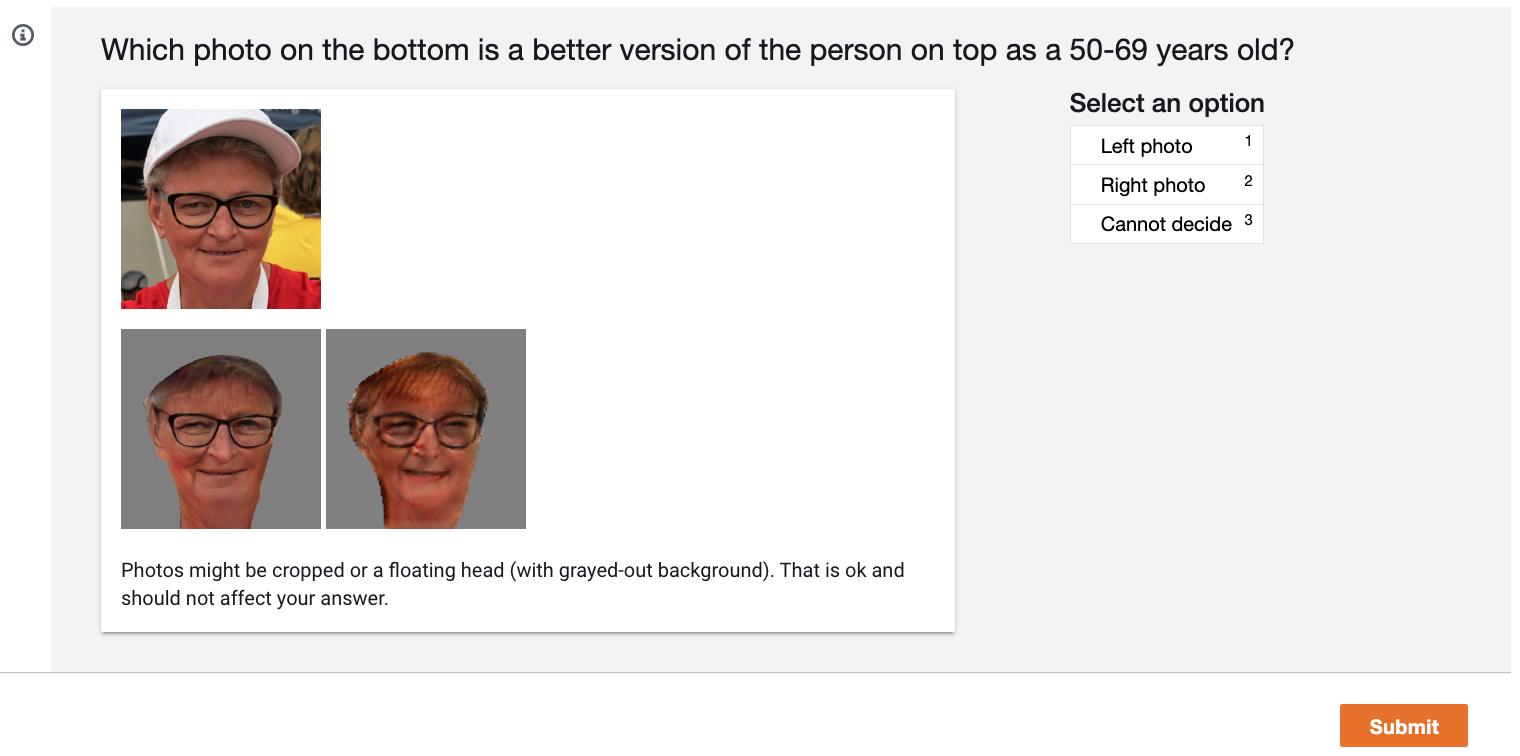}
    \caption{User study interface. We asked 3 different questions to assess age, identity and overall quality.}
    \label{fig:user_study_ui}
\end{figure}

\begin{table}[t!]
\begin{minipage}[t]{\textwidth}
\centering
\fontsize{6}{8}\selectfont
\setlength{\tabcolsep}{3pt}
\begin{tabular}{l@{\hskip 0.4cm}ccccccccccccccc}
\toprule
Age range:     & \multicolumn{2}{c}{0--2} & \multicolumn{2}{c}{3--6}  & \multicolumn{2}{c}{7--9} & \multicolumn{2}{c}{15--19} & \multicolumn{2}{c}{30--39}  & \multicolumn{2}{c}{50--69} & & \multicolumn{2}{c}{All}      \\
\cmidrule(l{2.5pt}r{2.5pt}){2-3} \cmidrule(l{2.5pt}r{2.5pt}){4-5} \cmidrule(l{2.5pt}r{2.5pt}){6-7} \cmidrule(l{2.5pt}r{2.5pt}){8-9} \cmidrule(l{2.5pt}r{2.5pt}){10-11} \cmidrule(l{2.5pt}r{2.5pt}){12-13} \cmidrule(l{2.5pt}r{2.5pt}){15-16}
               & \cite{wang2018face} & Ours              & \cite{wang2018face}      & Ours          & \cite{wang2018face}      & Ours         & \cite{wang2018face}       & Ours          & \cite{wang2018face}       & Ours          & \cite{wang2018face}       & Ours          & & \cite{wang2018face}       & Ours  \\
\midrule
Same identity $\uparrow$  & 14 & \textbf{20} & 19 & \textbf{23} & 24 & 24 & 20 & \textbf{25} & \textbf{24} & 22 & 19 & \textbf{23} & & 120 & \textbf{137} \\
Age difference $\downarrow$ & \textbf{1.0} & 3.4 & \textbf{2.1} & 3.2 & \textbf{4.5} & 5.1 & \textbf{6.4} & 10.3 & 8.2 & \textbf{7.4} & 13.3 & \textbf{6.5} & & \textbf{5.9} & 6.0 \\
\addlinespace[0.3em]
Overall better $\uparrow$ & 2 & \textbf{23} & 1 & \textbf{24} & 1 & \textbf{23} & 2 & \textbf{23} & 1 & \textbf{24} & 0 & \textbf{25} & & 7 & \textbf{142} \\
\bottomrule
\end{tabular}
\caption{User study results vs. IPCGAN~\cite{wang2018face} that was retrained on our dataset. Our results are better at preserving subject identity, and the two methods are extremely close at age accuracy. Most importantly, when asked which result is better overall, users preferred our results in 95\% of the cases (142 out of 150, compared to 7 for IPCGAN and 1 indecisive).}
\label{tbl:ipcgan_retrained}
\end{minipage} \\

\begin{minipage}[t!]{\textwidth}
\centering
\fontsize{6}{8}\selectfont
\setlength{\tabcolsep}{3pt}
\begin{tabular}{l@{\hskip 0.4cm}ccccccccccccccc}
\toprule
Age range:     & \multicolumn{2}{c}{0--2} & \multicolumn{2}{c}{3--6}  & \multicolumn{2}{c}{7--9} & \multicolumn{2}{c}{15--19} & \multicolumn{2}{c}{30--39}  & \multicolumn{2}{c}{50--69} & & \multicolumn{2}{c}{All}      \\
\cmidrule(l{2.5pt}r{2.5pt}){2-3} \cmidrule(l{2.5pt}r{2.5pt}){4-5} \cmidrule(l{2.5pt}r{2.5pt}){6-7} \cmidrule(l{2.5pt}r{2.5pt}){8-9} \cmidrule(l{2.5pt}r{2.5pt}){10-11} \cmidrule(l{2.5pt}r{2.5pt}){12-13} \cmidrule(l{2.5pt}r{2.5pt}){15-16}
               & \cite{liu2019stgan} & Ours              & \cite{liu2019stgan}      & Ours          & \cite{liu2019stgan}      & Ours         & \cite{liu2019stgan}       & Ours          & \cite{liu2019stgan}       & Ours          & \cite{liu2019stgan}       & Ours          & & \cite{liu2019stgan}       & Ours  \\
\midrule
Same identity $\uparrow$    & 16 & \textbf{22} & 24 & \textbf{25} & 25 & 25 & \textbf{25} & 24 & \textbf{25} & 24 & \textbf{25} & 24 & & 140 & \textbf{144} \\
Age difference $\downarrow$ & \textbf{4.0} & 4.4 & 15.7 & \textbf{6.2} & 19.8 & \textbf{9.5} & 17.5 & \textbf{12.3} & 13.3 & \textbf{7.0} & 23.1 & \textbf{7.7} &  & 15.6 & \textbf{7.8} \\
\addlinespace[0.3em]
Overall better $\uparrow$   & 5 & \textbf{20} & 6 & \textbf{18} & 3 & \textbf{20} & 3 & \textbf{20} & 3 & \textbf{21} & 1 & \textbf{24} & & 21 & \textbf{123} \\
\bottomrule
\end{tabular}
\caption{User study results vs. STGAN~\cite{liu2019stgan} that was retrained on our dataset. Our results are better at preserving subject identity, and have better age accuracy. Most importantly, when asked which result is better overall, users preferred our results in 82\% of the cases (123 out of 150, compared to 21 for STGAN and 6 indecisive).}
\label{tbl:stgan_retrained}
\ifarxiv
\vspace{-0.7cm}
\fi
\end{minipage}
\end{table}

The user interface of our user studies is presented in \Cref{fig:user_study_ui}. The same UI was used both for the studies in the main paper and in this supplemental document. In addition to the main paper user studies, we also wanted to verify that our results are not solely due to a better dataset. To this end, we retrained IPCGAN~\cite{wang2018face} and STGAN~\cite{liu2019stgan} on our data. 

In the following studies, we evaluate the results of 25 randomly selected photos on each of the 6 age classes, repeating each question 3 times, for a total of 2250 individual answers per user study.
Note that in these studies we can compare all 6 age groups, whereas in our other user studies we were limited by the choice to use the authors' pre-trained models which were not available for all ages.

User study results are in \Cref{tbl:ipcgan_retrained,tbl:stgan_retrained}. Indeed, we see that even when retrained on our data, there is a significant performance gap between our results and previous works \cite{liu2019stgan,wang2018face}. Our results are better at identity preservation, and either better or on-par in age accuracy. As explained in the main text, since overall quality is determined by both these factors and others such as image quality, we asked users which result is better overall. Our results were selected as better in 82\% (vs. StGAN) and 95\% (vs. IPCGAN) of the cases.

\begin{figure}[t]
\centering

\includegraphics[width=\textwidth]{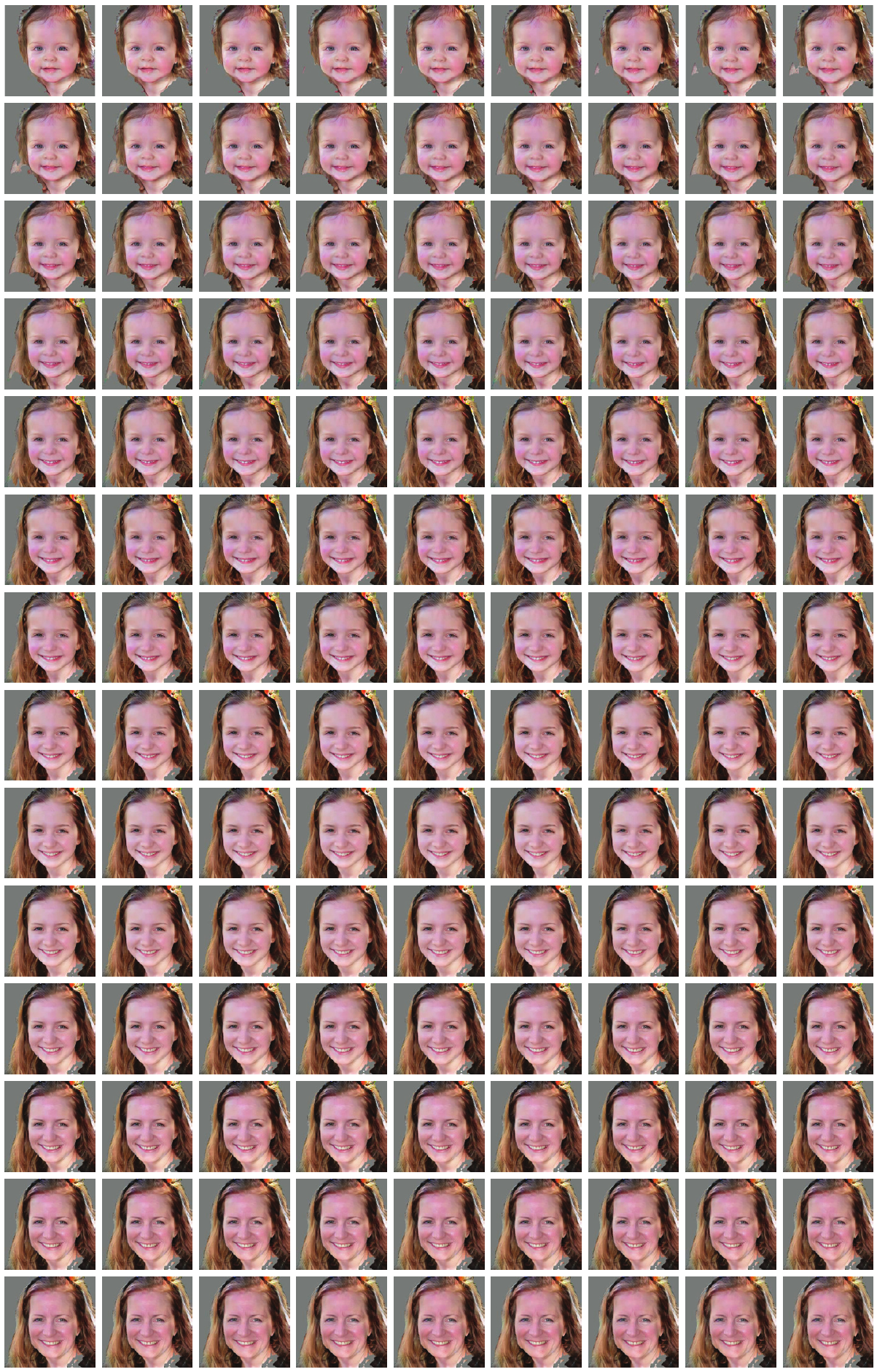}
\caption{Full lifespan transformation. Also see supplemental videos.}
\label{fig:video_frames_69222}
\end{figure}
\begin{figure}[t]
\centering

\includegraphics[width=\textwidth]{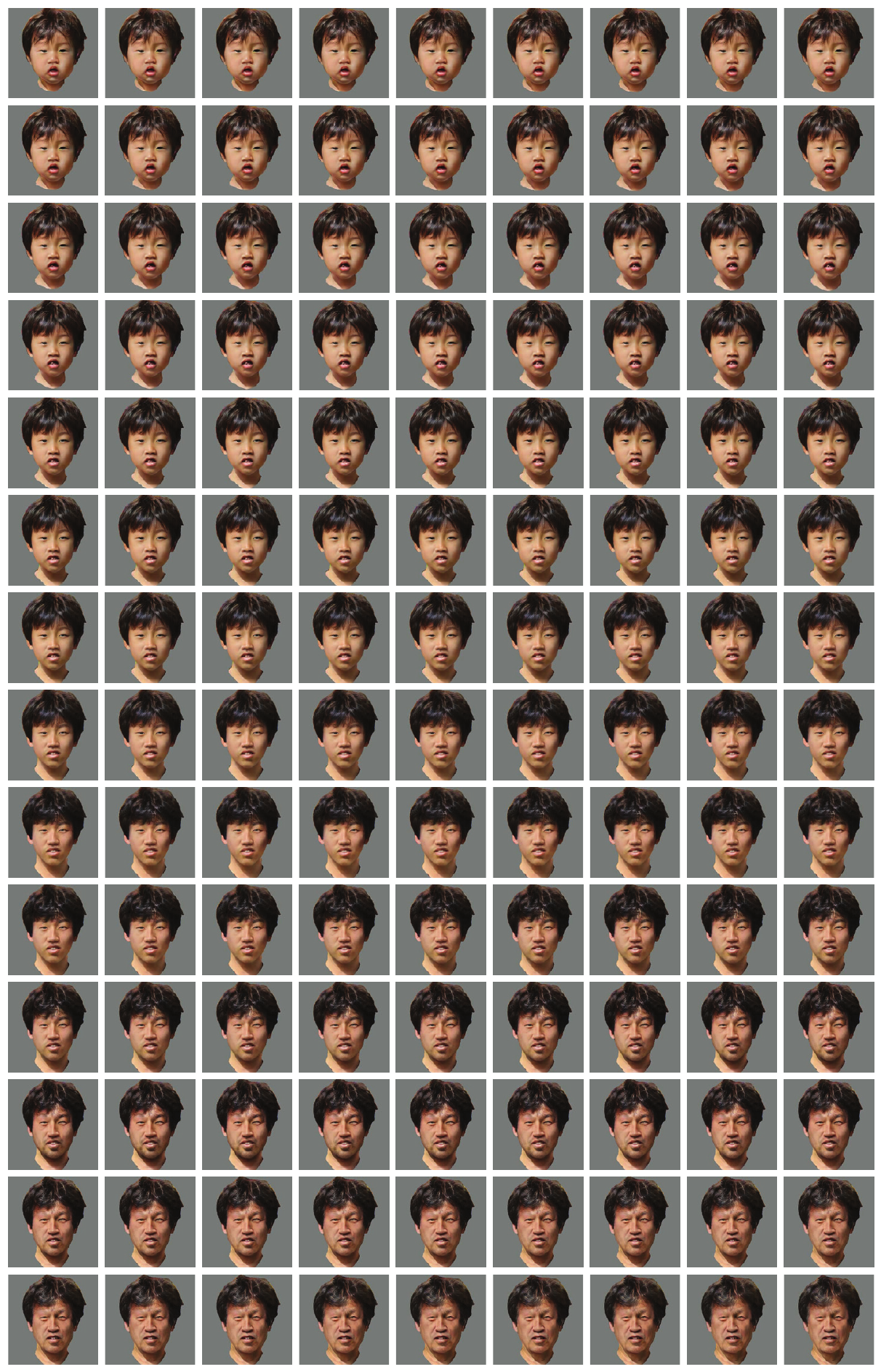}
\caption{Full lifespan transformation. Also see supplemental videos.}
\label{fig:video_frames_69235}
\end{figure}

\fi
\end{document}